\documentclass[journal,transmag]{IEEEtran}
\usepackage{graphicx}
\usepackage{graphics}
\usepackage{subfigure}
\usepackage{calligra}
\ifCLASSINFOpdf
\else
\fi
\begin{document}
%
\title{How deep learning works --- The geometry of deep learning}



%
%

\author{\IEEEauthorblockN{Xiao Dong, Jiasong Wu, Ling Zhou}
\IEEEauthorblockA{Faculty of Computer Science and Engineering, Southeast University, Nanjing, China}}

%



\IEEEtitleabstractindextext{%
\begin{abstract}
Why and how that deep learning works well on different tasks remains a mystery from a theoretical perspective. In this paper we draw a geometric picture of the deep learning system by finding its analogies with two existing geometric structures, the geometry of quantum computations and the geometry of the diffeomorphic template matching. In this framework, we give the geometric structures of different deep learning systems including convolutional neural networks, residual networks, recursive neural networks, recurrent neural networks and the equilibrium prapagation framework. We can also analysis the relationship between the geometrical structures and their performance of different networks in an algorithmic level so that the geometric framework may guide the design of the structures and algorithms of deep learning systems.
\end{abstract}

}

\maketitle



\IEEEdisplaynontitleabstractindextext

%
\IEEEpeerreviewmaketitle

\section{Introduction}
In the last decade, deep learning systems show a fascinating performance in solving different complex tasks. We have designed different system structures for different problems and revealed some general rules for the designing of deep learning systems. There are also theoretical attempts to understand deep learning systems from both mathematical and physical perspectives\cite{Lin_learningwork}. But still we are lacking of a theoretical framework to answer the question, why and how deep learning systems works. Also it's  highly desired that a theoretical framework of deep learning systems can be used to guide the design the structures of deep learning systems from an algorithmic level.

In this paper we try to fill this gap by drawing a geometric picture of deep learning systems. We build our geometric framework to understand deep learning systems by comparing the deep learning system with other two existing matural geometric structures, the geometry of quantum computations and the geometry of diffeomorphic template matching in the field of computational anatomy. We show that deep learning systems can be formulated in a geometric language, by which we can draw geometric pictures of different deep learning systems including convolutional neural networks, residual networks, recursive neural networks, fractal neural networks and recurrent neural networks. What's more, these geometric pictures can be used to analysis the performance of different deep learning structures and provide guidance to the design of deep learning systems from an algorithmic level.

The rest of this paper is arranged as follows. We will first give a brief overview of the geometry of quantum computations and the geometry of diffeomorphic template matching. Then we will explain the geometric framework of deep learning systems and apply our framework to draw correspondent geometric pictures of different deep learning networks. Finally we will give a general optimization based framework of deep learning systems, which can be used to address the equilibrium propagation algorithm.

\section{What can geometry teach us}

It's well known that geometry is not only a core concept of mathematics, but also it plays a key role in modern physics. The great success of geometrization of physics tells us that the soul of physical systems lies in their geometric structures. It's natural to ask if geometry can help to reveal the secret of our human intelligence and our state-of-the-art artificial intelligence systems, deep learning systems. The answer is YES.

We will first introduce two interesting geometric structures in the fields of quantum computation and computational anatomy. We will see, their geometric structures share some similarities with deep learning systems and the geometric framework of deep learning systems can be built based on an understanding of these structures.

\subsection{Geometry of quantum computation}
 Geometry concepts have been widely discussed in formulating quantum mechanics and quantum information processing systems, including the geometry of quantum states and their evolution\cite{Ashtekar_geometryquantum}\cite{Heydari_geometryquantum}\cite{Andersson_geo_mixed_states}, the geometry of entanglement\cite{Levay_entanglement_connection}\cite{Levay_entanglement_twistor} and also the relationship between the spacetime structure and the geometry of quantum states\cite{Van_Raamsdonk_spacetime_entanglement}\cite{Stanford_complexity}\cite{Susskind_ER_bridge_nowhere}.

 In \cite{Nielsen_geometry}\cite{Nielsen_geometry2} a geometric framework was proposed for the complexity of quantum computations. Its basic idea is to introduce a Riemannian metric to the space of n-qubit unitary operators so that the quantum computation complexity becomes a geometric concept as given by the slogan \emph{quantum computation as free falling}. The key components of this framework can be summarized as follows.

\begin{itemize}
  \item An algorithm of a n-qubit quantum computation system is a unitary operation $U \in U(2^{n})$, which can evolve the n-qubit initial state $|\psi_{ini}\rangle=|00...0\rangle$ to the final state $|\psi_{fin}\rangle=U|\psi\rangle_{ini}$. $U(2^n)$ is the space of the unitary operations of n-qubits, which is both a manifold and a Lie group.

  \item Any physical realization of the algorithm $U$ is a curve $U(t)\in U(2^n),t\in[0,1]$ with $U(0)=I,U(1)=U$, where $I$ is the identity operator. A smooth curve $U(t)$ can be achieved by a Hamiltonian $H(t)$ so that $\dot{U}(t)=-iH(t)U(t)$, where $H(t)$ is in the tangent space of $U(2^n)$ at $U(t)$ and also an element of the Lie algebra $u(2^n)$.

  \item A Riemannian metric $\langle\cdot,\cdot\rangle_U$ can be introduced to the manifold $U(2^n)$ so that a Riemannian structure can be built on $U(2^n)$. The metric is defined as
  \begin{equation}\label{eq1}
     \langle H,J\rangle_U=[tr(H\mathcal{P}(J))+q tr(H\mathcal{Q}(J))]/2^n
  \end{equation}

      where $J,P$ are tangent vectors in the tangent space at $U$, and $\mathcal{P}(J)$ is the projection of $J$ to a simple local Hamiltonian space $P$, $\mathcal{Q}=1-\mathcal{P}$ and $q>>1$. With this metric any finite length curve $U(t)$ connecting the identity $I$ and $U$ will only contain local Hamiltonian when $q\rightarrow\infty$. Usually the geodesic $U_{geo}(t)$ connecting $I$ and $U$ has the minimal length, which is defined as the complexity of the algorithm $U$. We also know the geodesic is given by the EPDiff equation, which is obtained by solving the optimization problem $\min_{H(t)} \int\langle H(t),H(t)\rangle_{U}^{1/2}dt$.

  \item The quantum circuit model of quantum computations tells us that any $U$ can be approximated with any accuracy with a small universal gate set, which contains only simple local operators, for example only a finite number of 1 and 2 qubit operators. The quantum circuit model can then be understood as approximating $U(t)$ with a piecewise smooth curve, where each segment corresponds to a universal gate. It's proven in \cite{Nielsen_geometry}\cite{Nielsen_geometry2} that the number of universal gates to approximate $U$ within some constant error is given by $O(n^{t}d(I,U)^{3})$, where $d(I,U)$ is the geodesic distance between $I$ and $U$. So an algorithm $U$ can be efficiently realized by the quantum circuit model only when its geodesic distance to the identity operator $I$ is polynomial with respect to $n$.

 \item The set of operators that can be reached by a geodesic with a length that is polynomial with the qubit number n is only a small subset of $U(2^n)$. Equivalently, starting from a simple initial state $|\psi\rangle_{ini}=|000...0\rangle$, the quantum state that can be obtained efficiently by local quantum operations is a small subset of the quantum state space. This defines the complexity of a quantum state.

  \item Finding the geodesic $U^{geo}(t)$ can be achieved by either the lifted Jacobi field or standard geodesic shooting algorithm. The performance of the optimization procedure to find $U_{geo}(t)$ highly depends on the curvature of the Riemannian manifold. For the metric given in (\ref{eq1}), the sectional curvature is almost negative everywhere when $q$ is large enough. This means the geodesic is unstable and it's general difficult to find a geodesic.
\end{itemize}

\begin{figure}
  \centering
  \includegraphics[width=7cm]{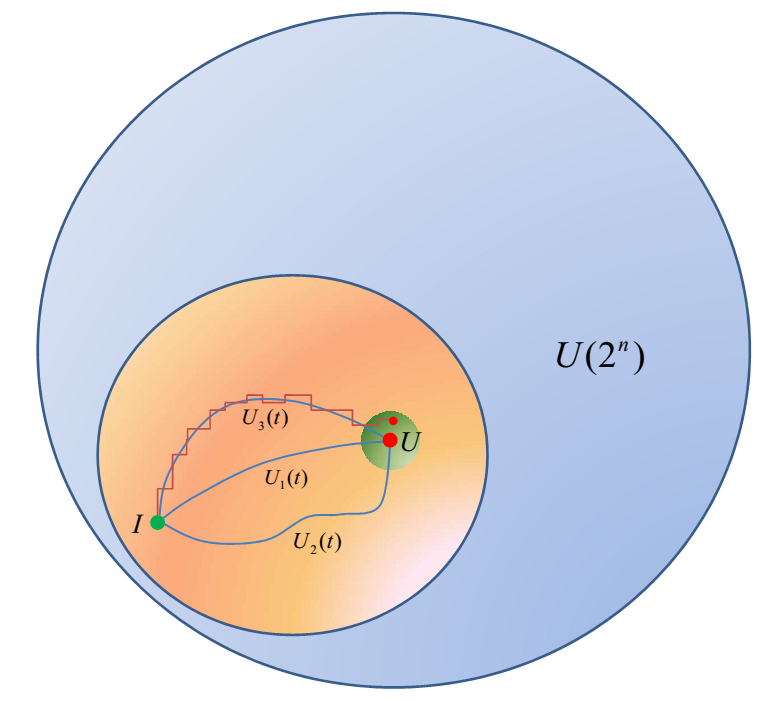}
  \caption{Geometry of the quantum computation. The yellow area gives the space of quantum algorithms that can be efficiently achieved by the quantum circuit model. Given a Riemannian structure on $U(2^n)$, there exist an arbitrary number of ways to achieve an algorithm $U$ and its complexity is given by the length of the geodesic connecting $I$ and $U$. Any efficient realization curve $U(t)$ can approximated by an universal gate set with a polynomial cost as shown by the red segmented line.}\label{fig1}
\end{figure}

The key lesson of the geometry of quantum computations is that:
\textbf{Under a local Hamiltonian preferred metric, the complexity of a general quantum algorithm $U$ is exponential with the number of qubit n. If the complexity $U$ is polynomial, it can be achieved efficiently with local operators, either by a continuous local Hamiltonian $H(t)$ or a set of universal local operators. The performance of the optimization procedure to find a good realization of $U$ depends on the curvature of the Riemannian manifold}.

We will see that this observation is closely related with the question of\emph{why and how cheap deep learning works}.

\subsection{Geometry of diffeomorphic template matching}
The diffeomorphic framework of computational anatomy aims to achieve a diffeomorphic matching between images on a Riemannian structure or a Lie group. The geometry of diffeomorphic template matching not only provides a mathematical framework for image registration but also validates the possibility of statistical analysis of anatomical structures on manifolds\cite{Beg2005_LDDMM}\cite{Hernandez2006_SVF}\cite{Lorenzi2011_OneParameterGroup}\cite{Durrleman2009_SpatiotemporalAtlas}\cite{Bruveris2011_MomentumMap}\cite{Holm2013_GeometryRegistration}\cite{Holm2009_EP_Metamorphosis}.

We can summarize the geometry of diffeomorphic template matching as follows:
\begin{itemize}
  \item The Riemannian geometry of diffeomorphic template matching starts with the claim that a diffeomorphic matching between two images $I_0$ and $I_1$ defined on a space $M=R^n$ ($n=2$ for 2 dimensional images or $n=3$ for 3 dimensional volume data) is achieved by a diffeomorphic transform $g \in Diff(M)$. $Diff(M)$ the diffeomorphism transformation group of the space $M$.

  \item A matching between $I_0$ and $I_1$ is achieved by a smooth curve $g(t)$ on $Diff(M)$ connecting the identity transform $I$ and $g$. The optimal $g(t)$ is obtained by an optimization problem
      \begin{equation}\label{eq1-1}
        \min_{u(t)}E=\min_{u(t)}\int_0^1 \langle u(t),u(t)\rangle_v dt+\frac{1}{\sigma^2}\|I_1-I_0\circ g_{1}\|_{L^2}^2
      \end{equation}
      where $\dot{g(t)}=u(t)\circ g(t)$, $g(0)=I$ and $\langle u(t),u(t)\rangle_v=\langle Lu(t),Lu(t)\rangle_{L^{2}}$ is a metric on the tangent space of the manifold $Diff(M)$ with $L$ a linear operator.

      This is the LDDMM framework\cite{Beg2005_LDDMM} and the optimization results in the Euler-Lagrange equation given by
      \begin{equation}\label{eq1-3}
       \nabla_{u(t)}E=2u_{t}-K(\frac{2}{\sigma^2}|Dg_{t,1}^{u_t}|\nabla J_{t}^{0}(J_{t}^{0}-J_{t}^{1}))=0
      \end{equation}
      where $J_{t}^{0}=I_{0}\circ g_{t,0}^{u_t}$,$J_{t}^{1}=I_{1}\circ g_{t,1}^{u_t}$ and $g_{s,t}^{u_t}$ is the transformation generated by $u(t)$ during the time internal $s$ to $t$. $K$ is the operator defined by $\langle a,b\rangle_{L^2}=\langle Ka,b\rangle_{v}$, i.e., $K(L^{+}L)=1$. In computational anatomy $K$ is usually taken as a Gaussian kernel, whose effect is to generate a smooth velocity field $u(t)$ that leads to a smooth transformation $g(t)$.

      The geometric picture of LDDMM is to find a geodesic $g(t)$ that transforms $I_0=I_o\circ g(0)$ to a near neighbour of $I_1$ so that $I_1\approx I_0\circ g(1)$. The optimal velocity field $u(t)$ is obtained with a gradient descent algorithm.

  \item An alternative approach is the geodesic shooting algorithm\cite{Vialard2012_GS_Atlas}\cite{Vialard2011_GS_Adjoint}\cite{Vialard2011_GS_KarcherMean}, which is essentially a variational problem on the initial velocity $u(0)$ of the geodesic $g(t)$ and the EPDiff equation is taken as a constraint.

  \item The sectional curvature of the volume preserving diffeomorphisms of the flat torus with a weak $L^2$ right invariant metric is negative in may directions so the geodesic on this manifold can be unstable. We indicate this point to show that here we might also face a similar negative curvature case as in the geometry of the quantum computation.

  \item If instead we do not introduce the metric $\langle u(t),u(t)\rangle_v$ in $Diff(M)$, $Diff(M)$ can be regarded as a Lie group. Then an alternative matching curve $g(t)$ can be formulated as a Lie group exponential map $exp(\mathbf{g}t),t \in [0,1]$ and $\mathbf{g} \in diff(M)$ with $diff(M)$ as the Lie algebra of $Diff(M)$. This is how the SVF framework\cite{Hernandez2006_SVF}\cite{Lorenzi2011_OneParameterGroup} formulates the template matching problem.

  \item There exists a mathematically strict framework for the geometry of template matching which shares lots of similarities with mechanical systems. For more information of the geometry of template matching, please refer to \cite{Lorenzi2011_OneParameterGroup}\cite{Bruveris2011_MomentumMap}\cite{Holm2013_GeometryRegistration}\cite{Holm2009_EP_Metamorphosis}\cite{Postnikov2001_RiemannianGeometry}.
\end{itemize}

\begin{figure}
  \centering
  \includegraphics[width=8cm]{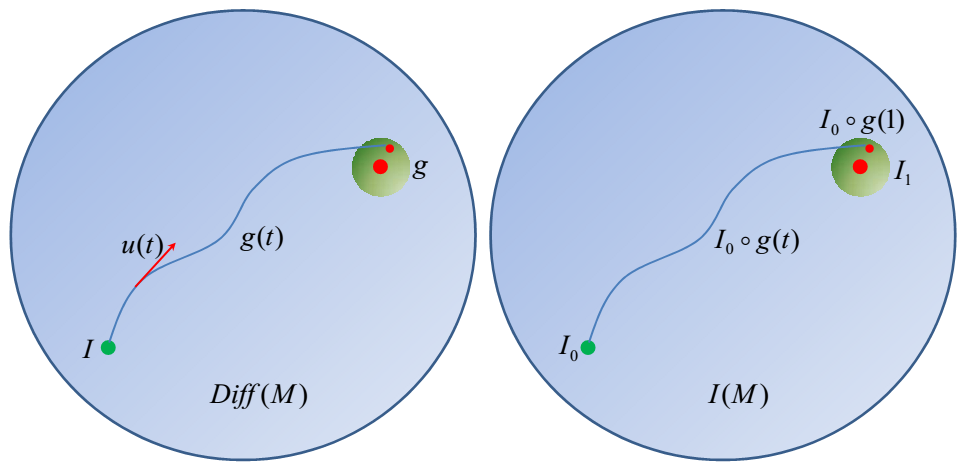}
  \caption{The geometry of the diffeomorphic template matching. Left: A diffeomorphic matching is a curve $g(t)$ in $Diff(M)$, the space of the diffeomorphic transformationg of the image; Right: The image defined on $M$ is smoothly deformed from $I_{0}$ to a near neighbour of $I_1$ by the curve $g(t)$.}\label{fig2}
\end{figure}

\begin{figure}
  \centering
  \includegraphics[width=8cm]{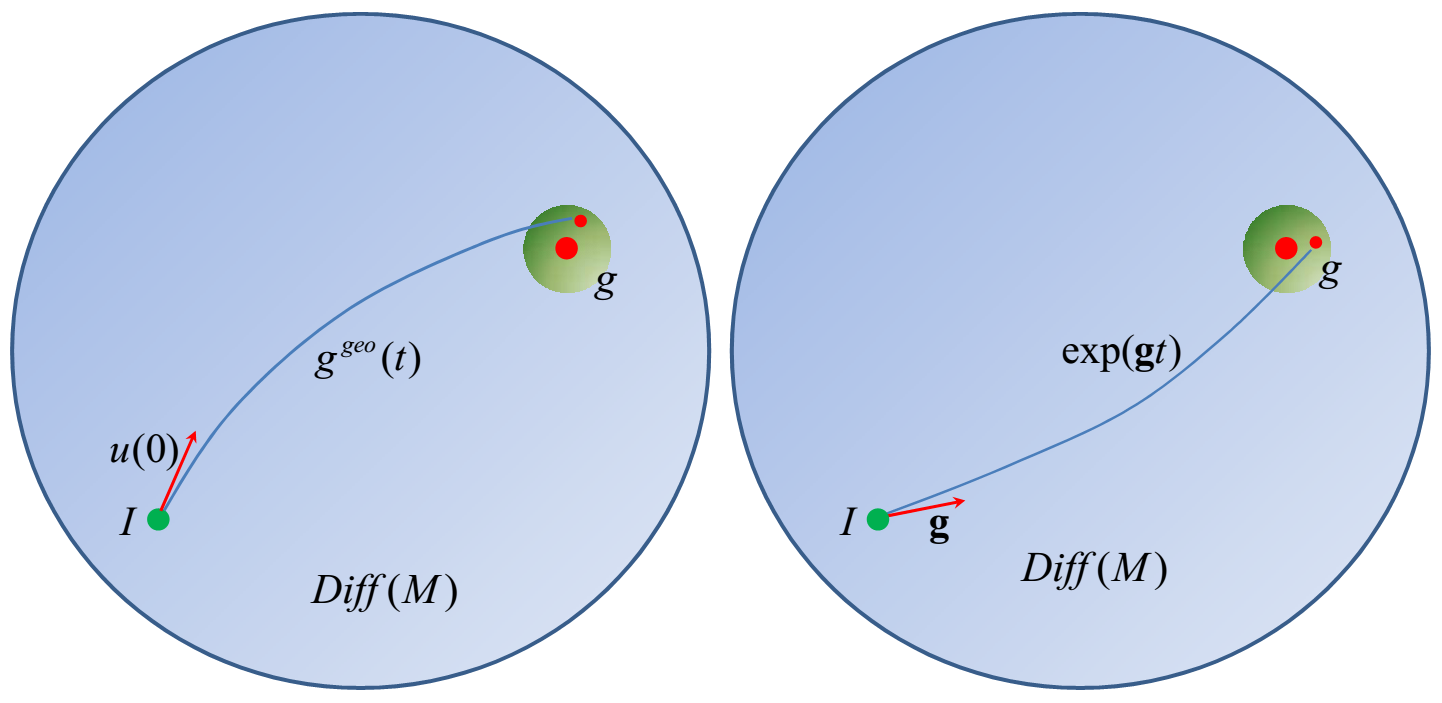}
  \caption{The geometry of the diffeomorphic computational anatomy. Left: The geometry of the template matching on Riemannian manifolds, where the optimal matching is achieved by a geodesic; Right: The geometry of the template matching on a Lie group, where the optimal matching is given by the Lie group exponential mapping.}\label{fig3}
\end{figure}

The key concept of the geometry of template matching is:
\textbf{The template matching can be achieved by either a Riemannian geodesic or a Lie group exponential mapping in $Diff(M)$ that transforms the source image to approximate the destination image. Finding the optimal curve can be formulated as an optimization problem. The Riemannian metric or the Lie group structure of $Diff(M)$ play a central role in the optimization procedure.}

Obviously the two geometric structures share lots of common characteristics. \textbf{Both systems consider the geometry of a mapping $T \times V\rightarrow V$, where $V$ is a vector space (the state space of n-qubits or the space of images on $M=R^{n}$ respectively) and $T$ is the space of transformations on $V$ ($SU(2^n)$ or $Diff(M)$). By introducing a Riemannian metric on $T$ and an inner product on $V$, both of them address the problem of finding a curve on a Riemannian manifold ($SU(2^n)$ or $Diff(M)$) to connect the identity operator with a target operator($U$ or $g$). Equivalently this curve can achieve a transformation from an initial state ($|\psi\rangle_{ini}$ or $I_0$) to a near neighbourhood (defined by the inner product on $V$) of a final state ($|\psi\rangle_{fin}$ or $I_1$) in $V$. The problem can be formalized as an optimization task and the optimal curve is a Riemannian geodesic (an Lie group exponential for SVF). The stability of geodesics and the convergence property of the optimization procedure highly depend on the Riemannian metric of the Riemannian manifold. In both geometric structure, we have to fact the negative sectional curvature problem.}

Besides their similarities, there exist a key difference between these two geometric structures. The geometry of the quantum computation addresses the problem to reach a point efficiently by local operations. So it focuses on the computability of a physical system. For more discussion on the computability and physical systems, please refer to \cite{Nielsen_geometry}. On the contrary, in the template matching problem, we do not emphasis to use local operations. The reason falls in that the deformation curve $g(t)$ is in fact generated purely by local operations since it's an integration curve of a time-varying velocity field, which is essentially local. Also in the template matching problem, it seems that we do not care if a transformation $g \in Diff(M)$ can be efficiently computed.

The reason for the difference between them is related to the concept of the diameter of a Riemannian manifold, which is defined by the maximal geodesic distance between arbitrary points on the manifold. So for the geometry of the quantum computation, the manifold has an exponentially large diameter with respect to the qubit number $n$. So only a subset of the operations can be efficiently carried out. For the template matching problem, the diameter is usually finite. In fact the real situation is quite complex. For example, if we work on the volume-preserving diffeomorphism group $SDiff(M^n)$ of a n-dimensional cube $M^n$ in $R^n$, the diameter is finite for $n\geq 3$ but infinite for $n=2$ if a $L^2$ right-invariant metric is used. For $Diff(M^n)$, a $H^1$ metric results in a finite diameter, and a $L^2$ metric can even lead to a vanishing geodesic distance, which means any two points can be connected with a geodesic with an arbitrary small distance. So generally we assume all the transformations in the template matching problem can be efficiently computed.

We will see that the diameter of the Riemannian manifold of the deep learning systems is closed related with the question \emph{why deep learning works}.

\section{Geometry of Deep Learning}
Based on the above mentioned geometric structures, we propose a geometric framework to understand deep learning systems. We will first focus on the most popular deep learning structure, the deep convolution neural network(CNN). Then the same framework can be used to draw geometric pictures of other systems including the Residual Network(ResNet), the recursive neural network, the fractal neural network and the recurrent network.

\subsection{Geometry of convolutional neural networks}
The convolutional neural network is the most well studied deep learning structure, which can achieve reliable object recognition in computer vision systems by stacking multiple CONV-ReLU-Pooling operations followed by a fully connected network. From a geometric point of view, CNN achieves the object classification by performing a nonlinear transformation $U_{CNN}$ from the input image space, where the object classification is difficult, to the output vector space, in which the classification is much easier. We will show that there exists an exact correspondence between CNN and the geometry of the quantum circuit model of quantum computation, where $U_{CNN}$ corresponds to the unitary operator $U$ of the quantum computation and the CONV-ReLU-Pooling operation plays the role of local universal gates.

The geometric structure of CNN is constructed as follows.
\begin{itemize}
  \item \textbf{The manifold}\\ We first construct the mapping $T_{CNN} \times V_{CNN}\rightarrow V_{CNN}$. In a CNN with $L$-layers, each layer accomplishes a transformation of the input data. We can easily embed the input and output data of each step into a Euclidean space $R^n$ with a high enough dimension $n$. We call it $V_{CNN}$ and its dimension is determined by the structure of the CNN including the input data size and the kernel number of each level. Accordingly $T_{CNN}$ can be taken as the automorphism of $V_{CNN}$. The goal of CNN is then to find and realize a proper transformation $U_{CNN} \in T_{CNN}$ by constructing a curve $U_{CNN}(t)$ on $T_{CNN}$, which consists of L line segments, to reach $U_{CNN}$.

  \item \textbf{The Riemannian metric}\\ It's easy to see that the local CONV-ReLU-Pooling operation of each layer corresponds to the local unitary operations in the quantum computation system. The details of each CONV-ReLU-Pooling operation define the allowed local operations, i.e., the set $P$ in the quantum computation system.

      Now we have two different ways to understand the metric on $T_{CNN}$. We can define a similar metric as (\ref{eq1}), which means that at each layer of the CNN, this metric only assigns a finite length to the allowed local operations defined by the CONV-ReLU-Pooling operation of this layer. So in this picture the metric changes during the forward propagation along the CNN since the configuration of the CONV-ReLU-Pooling operator differs at each layer. So $U_{CMM}(t)$ is a curve on a manifold with a time-varying metric. To remedy our Riemannian structure from this complicated situation, we can define another computational complexity oriented metric by scaling the metric at each layer by the complexity of the CONV-ReLU-Pooling operation. For example, if at a certain layer we have $N_{k}$ kernels with a kernel size $K_x\times K_y \times K_z$ and the input data size is $S_x\times S_y \times S_z$, then we scale the metric by a factor $N_kK_xK_yK_zS_xS_yS_z$. With this metric, $U_{CNN}(t)$ is a curve on $T_{CNN}$ with this complexity oriented metric and the length of $U_{CNN}(t)$ roughly corresponds to the complexity of the CNN, if the nonlinear operations are omitted. Of course, the nonlinear ReLU and pooling operation in fact also change the metric. Existing research works showed that they have a strong influence on the metric as will be explained below.

  \item \textbf{The curvature}\\ If the CNN is formulated as a curve on a Riemannian manifold, then obviously the convergence of the training procedure to find the curve highly depends on the curvature of the manifold. It's difficult to compute explicitly the curvature of $T_{CNN}$. But we may have a reasonable guess since we know that the Riemannian manifold of the quantum computation, which has a very similar metric, has a curvature which is almost negative everywhere. Of course the metric of the CNN is more complex since here we also have the nonlinear component. If $T_{CNN}$ also has a almost negative curvature under our first definition of the metric shown above, then our second scaled metric will result in a highly curved manifold with a large negative curvature depending on the size of kernels and the selection of the activation function. Roughly speaking, a larger kernel size and a stronger nonlinearity of the activation function lead to a higher negative curvature.

   \item \textbf{The target transformation $U_{CNN}$}\\ Different with the case of the quantum computation, where the algorithm $U$ is known, here we do not have $U_{CNN}$. $U_{CNN}$ is found in a similar way as in the case of the template matching, where the target transformation $g$ is obtained by an optimization to minimize the error between the transformed source image and the destination image $\|I_1-I_0\circ g_{1}\|_{L^2}^2$. In CNNs, we find $U_{CNN}$ in a similar way. The only difference is that now the optimization is carried out on an ensemble of source and destination points, i.e., the training data with their corresponding labeling.

    \item \textbf{The curve $U_{CNN}(t)$}\\    $U_{CNN}(t)$ is a piece-wise smooth curve consisting of L line segments, where the length of each line segment corresponds to the computational complexity of each layer. So geometrically a CNN is to find a curve with a given geometrical structure to connect $I$ and $U_{CNN}$ on $T_{CNN}$.
\end{itemize}

The above given Riemannian is built on the transformation space $T_{CNN}$. In fact just like the cases in both the quantum computation and the template matching, equivalently we can have another Riemannian manifold on the data space $V_{CNN}$. We can regard the CNN achieves a change of the metric from the input data space to the final output data space defined by $\langle w,v\rangle_{in}=\langle U_*w,U_*v\rangle_{out}$, where $\langle \cdot,\cdot\rangle_{in}$ and $\langle \cdot,\cdot\rangle_{out}$ are the metric on the input and output manifolds respectively and $U_*$ is the differential of the transformation $U$. So in this construction, the metric of the data space changes along the data flow in the network. As we indicated before, the input data can not be easily classified in the input data space by defining a simple distance between points, but it's much easier in the output data space even with an Euclidean distance. This means by applying the transformation $U_{CNN}$, the CNN transforms a highly curved input data space to a flatter output data space by a curvature flow like mechanism. This gives an intuitive picture on how a complex problem can be solved with the CNN by flattening a highly curved manifold as in \cite{poole2016exponential}. But it's difficult to estimate how the curvature of the input data space except for some special cases as in \cite{Nielsen_geometry}. In the rest of this paper we will focus on the Riemannian structure on $T_{CNN}$ since it's relatively easier to analysis the curvature of $T_{CNN}$.

\begin{figure}
  \centering
  \includegraphics[width=8cm]{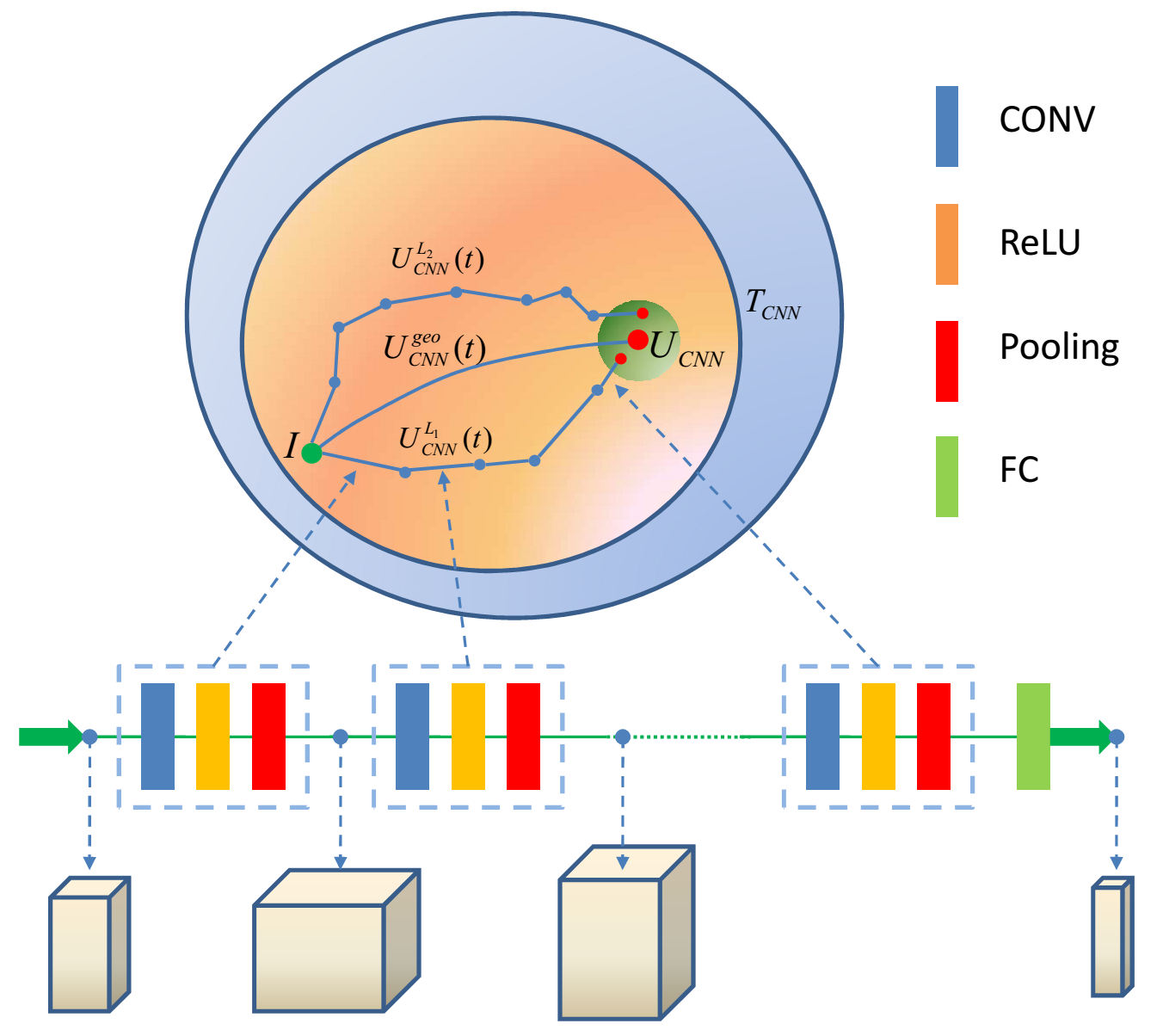}
  \caption{The geometry of CNNs. Similar with the geometry of quantum computation, $T_{CNN}$ is the space of all possible transformations that can be realized by a CNN. The Riemannian metric on $T_{CNN}$ is determined by the design of the CONV-ReLU-Pooling operations of the CNN and accordingly the curvature of the Riemannian manifold will influence the optimization procedure during the training phase of CNNs. The yellow area is the subspace of all the transformations that can be efficiently achieved by CNNs. Any given CNN structure corresponds a curve in $T_{CNN}$, where each line segment corresponds to the CONV-ReLU-Pooling operation of each layer and the length of the line segment is proportional to the computational complexity of this layer. Compared with the geometry of template matching, a CNN does not aim to find the optimal curve, i.e. the geodesic, instead it aims to find a curve with a fixed length $L$ which is proportional with the depth of a CNN.}\label{fig4}
\end{figure}

We now show how the design of the CNN structure influence the geometry and how the geometric structure can affect the performance of CNNs.

First we need to clarify how the components of CNNs will affect the geometric structure.
\begin{itemize}
  \item \textbf{Fully connected network}\\ Usually there is a fully connected network at the final stage of CNNs. From existing experimental works we know that its main function is to collect multiple modes of a certain category of object. For example, if the task is to identify cats, the FC network is in charge of collecting the information of cats with different locations and gestures. Also we know with the increase of the kernel number and depth, the FC part can be omitted. In another word, the convolutional network before the FC network in fact has already achieved an over classification of the objects, the FC network only achieve a clustering of the information. So the FC network does not play an important role in the geometry of CNN.

  \item \textbf{Depth of CNN}\\ The depth of a CNN is related to the total length of the curve $U_{CNN}(t)$, i.e., the complexity of the CNN. Obviously a deeper CNN can potentially achieve more complex transformation in $T_{CNN}$.

  \item \textbf{CONV operation}\\ The size and the number of kernel of the CONV operaion leads to different $V_{CNN}$ and $T_{CNN}$ as described above. Also as a component of the local CONV-ReLU-Pooling operation, CONV influences the complexity of local operators and the metric as well.  Generally a bigger kernel size and kernel number increase the complexity of allowed local operators. This can be understood if we compare it with the Riemannian metric defined in the geometry of quantum computations. Also the CONV of each layer changes the structure of the curve $U_{CNN}(t)$.

  \item \textbf{ReLU operation}\\ The role of ReLU operation (or a general activation function) falls in two folds: Firstly it provides the necessary nonlinear component to achieve the target nonlinear transformation $U_{CNN}$. Secondly different selections of the activation function show different nonlinear property, so it also influences the complexity of local operations and the Riemannian metric. The existing experimental works on the influence of the activation function on the convergence property of CNN indicate that it may have a very strong effect on the Riemannian metric, which leads to a change of the curvature of the Riemannian manifold $T_{CNN}$. This observation helps to analysis the influence of the activation function on the convergence of the training procedure.

   \item \textbf{Pooling operation}\\ The function of the pooling operation is also two folds: On one hand, it can reduce the complexity by changing  $V_{CNN}$ and $T_{CNN}$. On the other hand it also plays a role in constructing the metric, correspondent to the kernel $K$ in the LDDMM framework. By using pooling operations, we can improve the smoothness and robustness of the curve $U_{CNN}(t)$ since $U_{CNN}$ is estimated by a statistical optimization using samples from input image space, i.e., the training data. The pooling operation can be understood as an extrapolation of the samples so that the information of each sample is propagated to its near neighbourhood.
\end{itemize}

Given the above analysis, we can answer the following questions from a geometric point of view:
\begin{itemize}

  \item Why can CNNs work well in computer vision systems?\\
  There are some theoretical considerations to answer this question. In \cite{Lin_learningwork} it's claimed that the special structure of the image classification task enables the effectiveness of CNN, where the problem is formulated as the efficient approximation of a local Hamiltonian system with local operations of CNNs. By checking the analogy between the CNN and the quantum computation, we see that the argument of \cite{Lin_learningwork} is equivalent to say that for a task that can be solved by a deep CNN, its $U_{CNN}$ falls in the subset that can be achieved by local CONV-ReLU-Pooling operators, just as the subset of $U(2^n)$ that can be approximated by simple local universal gates in quantum computations. So the answer to this question is trivial, CNN works since the problem falls in the set of solvable problems of CNNs. A direct question is, can we design a task that does not belong to this solvable problem set? A possible design of such a task is to classify images that are segmented into small rectangular blocks and then the blocks are randomly permuted. Obviously such images are generated by global operations which can not be effectively approximated by local operations, and therefore the classification of such images can not be achieved efficiently by structures like CNN. So we can say, the diameter of the Riemannian manifold $T_{CNN}$ is exponentially large with respect to the size of the input data and only those problems which can be solved by a transformation $U_{CNN}$ that has a polynomially large geodesic distance to the identity transformation $I$ can be solved by CNNs.

  \item Can we justify if a problem can be efficiently solved by CNNs?
  Generally this is difficult since this is the same question as to ask if a unitary operator can be decomposed into a tensor product of simple local unitary operators, or to ask if a n-qubit quantum state has a polynomial state complexity with respect to $n$. We already know both of them are difficult questions.

  \item Why deep CNN?\\
Geometrically, training a CNN with a given deepth $L$ is equivalent to find a curve of length roughly proportional to $L$ in $T_{CNN}$ to reach the unknown transformation $U_{CNN}$. In deep learning systems, this is achieved by a gradient descent based optimization procedure, i.e., the training of CNNs. It should be noted that the CNN is not to find a geodesic, instead it aims to find a curve with a fixed geometric structure, i.e., a curve consisted of multiple line segments with given lengthes determined by the CNN structure. The reason to use the deep CNN is just that the length of $U_{CNN}(t)$ should be longer than the length of the geodesic connecting the identity transformation $I$ and $U_{CNN}$ on the Riemannian manifold $T_{CNN}$. Otherwise it will not work since a short $U_{CNN}(t)$ will never reach $U_{CNN}$. On the other side, it's not difficult to guess that a too deep CNN may also fail to converge if the curve $U_{CNN}(t)$ is too long due to both the depth $L$ and an improper design of CONV-ReLU-Pooling operations. This can be easily understood since to reach a point $U_{CNN}$ with a longer curve $U_{CNN}(t)$ is just to find an optima in a larger parameter space. Intuitively we prefer a design of CNNs that leads to a curve $U_{CNN}(t)$ with a length just above the geodesic distance between $I$ and $U_{CNN}$. To achieve this, we need a balance between the network depth and the complexity of CONV-ReLU-Pooling operations. This observation will play a key role in the analysis of ResNet below.

 \item Why does ReLU work better than the sigmoid in deep CNNs?\\
 A typical answer to this problem is that ReLU can improve the vanishing gradient problem so that a better performance can be achieved. Another idea is that ReLU can change the Fisher information metric to improve the convergence but the sigmoid function can not. Here we explain this from a complexity point of view. As mentioned before, the selection of the activation function will influence the complexity of the local operation. Obviously ReLU is a weaker nonlinear function than the sigmoid. This means for a given CNN structure, using ReLU will have a lower representation capability than using the sigmoid so that the CNN using ReLU has a smaller complexity. This is to say that the a CNN using a sigmoid need to search for an optima in a larger state space. It's obvious that as far as the complexity of the CNN is above the length of the geodesic, it's preferable to use a simpler CNN so that a better convergence can be achieved. It seems there exists a balance between the complexity of the CONV operation and the activation function so that the complexity of the complete system can be distributed equally along the network. With our first definition of the metric, the intuitive geometric picture is that the sigmoid function will introduce a metric so that the curvature of the Riemannian manifold will be more \emph{negative}, if our guess of the curvature of $T_{CNN}$ is true. A higher negative curvature of course will make the curve $U_{CNN}(t)$ more unstable so that a small change of the parameters of the CNN will result in a large deviation of the final point of $U_{CNN}(t)$. This corresponds to the observed vanishing gradient problem during the back-propagation procedure. So for a general deep CNN, the sigmoid is too complex and ReLU is a proper activation function that can produce a Riemannian manifold with a proper curvature.

\item How difficult is it to optimize the structure of CNNs?
By the optimization of the structure of CNNs, we mean to find the optimal CNN structure (for example the depth, the number and sizes of kernels of a CNN), which can accomplish the classification task with the minimal computational complexity. Geometrically this is to construct and define a metric on the Riemannian fold $T_{CNN}$, under with the length of a curve is proportional to the computational complexity of a CNN structure; and (b)realize the geodesic $U_{CNN}^{geo}$ that connects the identity operator $I$ with $U_{CNN}$. Generally this is difficult, especially when the curvature property of the Riemannian manifold is unknown. This is the same as to find the optimal realization of an algorithm with a given universal logic gate set. For the geometry of quantum computation, the curvature is almost negative so that it's difficult to achieve a general operator\cite{Nielsen_geometry}. If the manifold of the fundamental law, the quantum computation, has a negative curvature, it will not be surprising if we assume the geometry of deep learning has a similar property. This point seems to be partially confirmed by the recent work\cite{poole2016exponential}, where a manifold with an exponentially growing curvature with the depth of a CNN can be constructed. If a CNN is applied to such an input data manifold with an exponentially large \emph{negative} curvature, then the optimization will be difficult to converge.
\end{itemize}

\subsection{Geometry of residual networks}
ResNet is a super-deep network structure that shows superior performance than normal CNNs\cite{He2015_ResNet}\cite{He2016_ResNet2}. Given the geometry of CNNs, it's straight forward to draw the geometric picture of ResNets.

Similar with CNNs, ResNets also aim to find a curve $U_{ResNet}(t),t\in[0,1]$ to reach $U_{ResNet}$. The difference is that $U_{ResNet}(t)$ consists of a large number of short line segments corresponding to simple operations $U_{ResNet}(l\delta),l=1,2,...,L, L\delta=1$. It's easy to tell that when $L$ is big enough, $U_{ResNet}(l\delta)$ is near the identity operation $I$ so that the first-order approximation of $U_{ResNet}(l\delta)$ gives $U_{ResNet}(l\delta)\approx I+H_{ResNet}(l\delta)$, where $H_{ResNet}(l\delta)$ is a weak local nonlinear transformation. This gives the structure of the ResNet, i.e., each layer of the ResNet is the addition of the original and the residual information.

\begin{figure}
  \centering
  \includegraphics[width=8.5cm]{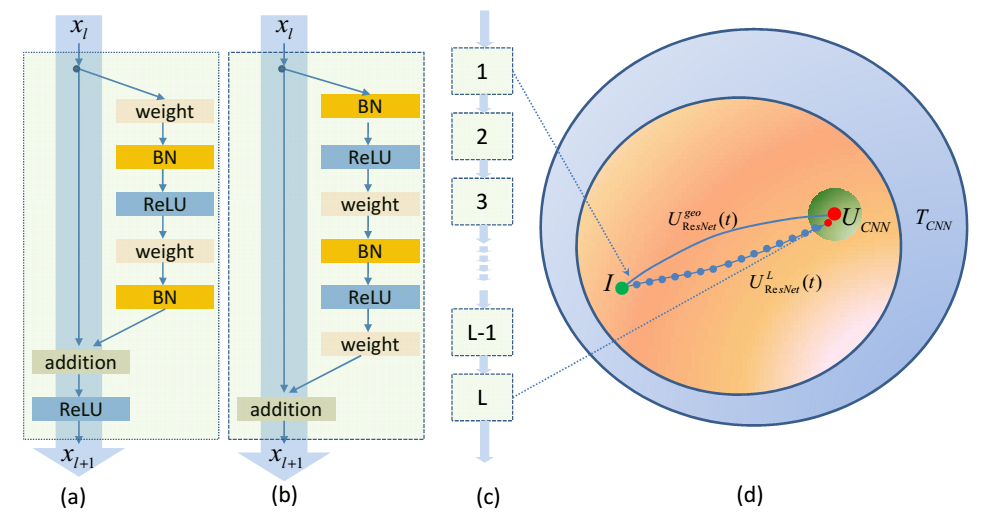}
  \caption{The geometry of ResNets. (a)The original residual unit of ResNets and (b) a new designed residual unit proposed in \cite{He2016_ResNet2}. (c) The deep ResNet built on residual units (a)(b). (d)ResNets hold a similar geometric picture as CNNs. The new feature of ResNet is that the curve $U_{ResNet}$ is a smooth curve consisted of a large number of simple near-identity transformations so that they can be efficiently approximated by the residual unit structure.}\label{fig5}
\end{figure}

The reason that ResNet works better falls in the following facts:
\begin{itemize}
  \item The ResNet achieves a roughly uniform distribution of the complexity of the problem along the curve $U_{ResNet}(t)$. Such a structure may lead to a smooth or stable information flow along the network and help to improve the convergence of the optimization procedure.
  \item The fact that $U_{ResNet}(l\delta)\approx I$ reduces the searching space of $H_{ResNet}(l\delta)$. For example, if the space $T_{ResNet}$ is $U(2^n)$, then $H_{ResNet}$ is in the Lie algebra $u(2^n)$, which has a smaller dimension than $U(2^n)$. This may improve the performance of the optimization procedure.
\end{itemize}

We can also use this framework to analysis the properties of ResNets observed in \cite{He2016_ResNet2}.
In \cite{He2016_ResNet2} the ResNet is formulated in a general form
\begin{eqnarray}
  y_l &=& h(x_l)+F(x_l,W_l) \\
  x_{l+1} &=& f(y_l)
\end{eqnarray}
where $x_l$ and $x_{l+1}$ are input and output of the $l$th-level and $F$ is a residual function. In the original ResNet, $h(x_l)=x_l$ and $f=ReLU$.

The main results of \cite{He2016_ResNet2} are:
\begin{itemize}
  \item If both the shortcut connection $h(x_l)$ and the after-addition activation $f_(y_l)$ are identity mappings, the signal could be directly propagated in both the forward and backward directions. Training procedure in general becomes ealier when the architecture is closer to these conditions.
  \item When $h(x_l)$ is selected far from the identity mapping, for example as multiplicative manipulations(scaling, gating, $1\times 1$ convolutions and dropout), the information propagations is hampered and this leads to optimization problems.
  \item The pre-activation of ReLU by putting the ReLU as part of $F$ and setting $f$ as the identity operation improves the performance. But the impact of $f=ReLU$ is less severe when the ResNet has fewer layers.
  \item The success of using ReLU in ResNet also confirms our conclusion that ReLU is a proper weak nonlinear function for deep networks and the sigmoid is too strong for deep networks.
\end{itemize}

From the geometric point of view, the above mentioned observation are really natural.

For a deep ResNet, $U_{ResNet}(l\delta)=I+H_{ResNet}(l\delta)$, this is exactly the case that both $f$ and $h$ are identical mappings. If $h$ is far from the identity mapping, then the difference between the identity mapping and $h$ need to be compensated by $F$, this will make $F$ more complex and leads to optimization problems.

The pre-activation also aims to set $f$ as the identity mapping and the weak nonlinear ReLU is absorbed in $F$. If the ResNet has fewer layers, then $U_{ResNet}(l\delta)$ is a little far from the identity mapping so that $U_{ResNet}(l\delta)=I+H_{ResNet}(l\delta)\approx ReLU(I+H'_{ResNet})$ is valid. This can explain the observation that the pre-activation is less crucial in a shallower ResNet.

The geometric picture easily confirms that the best ResNet structure is to replicate $U_{ResNet}(l\delta)=I+H_{ResNet}(l\delta)$. Another related observation is that the highway network\cite{Srivastava2015_highway} usually performs worse than the ResNet since it introduces unnecessary complexity to the system by adding extra nonlinear transformations to the system so the structure of the highway network does not match the natural structure given by the first-order approximation.

\subsection{Geometry of recursive neural networks}
 Recursive neural network is commonly used in the natural scene images or natural language processing tasks.

 Taking the recursive neural network for the structure prediction described in \cite{Socher_Recursive}\cite{Socher2013RecursiveDM} as an example, the goal is to learn a function $f:\mathcal{X}\rightarrow \mathcal{Y}$, where $\mathcal{Y}$ is the set of all possible binary parse trees. An input $x\in \mathcal{X}$ consists of (a) a set of activation vectors which represent input elements such as image segments or words of a sentence, and (b) a symmetric adjacency matrix to define which elements can be merged.

The recursive neural network accomplishes this task by finding the two sub mappings: (a) A mapping from the words of natural languages or segments of images to the semantic representation space. Combining this map with the adjacent matrix, the original natural input data is represented in a space ${\tilde{\mathcal{X}}}$; (b)A fixed rule to recursively parse the input data ${\tilde{\mathcal{X}}}$ to generate a parsing tree.

Geometrically we are interested in the following aspects:
\begin{itemize}
  \item What's the correspondent geometric picture of the fixed recursive rule?
  \item Why both the recursive rule and the representation mapping can be learned during training?
\end{itemize}

\begin{figure}
  \centering
  \includegraphics[width=8.5cm]{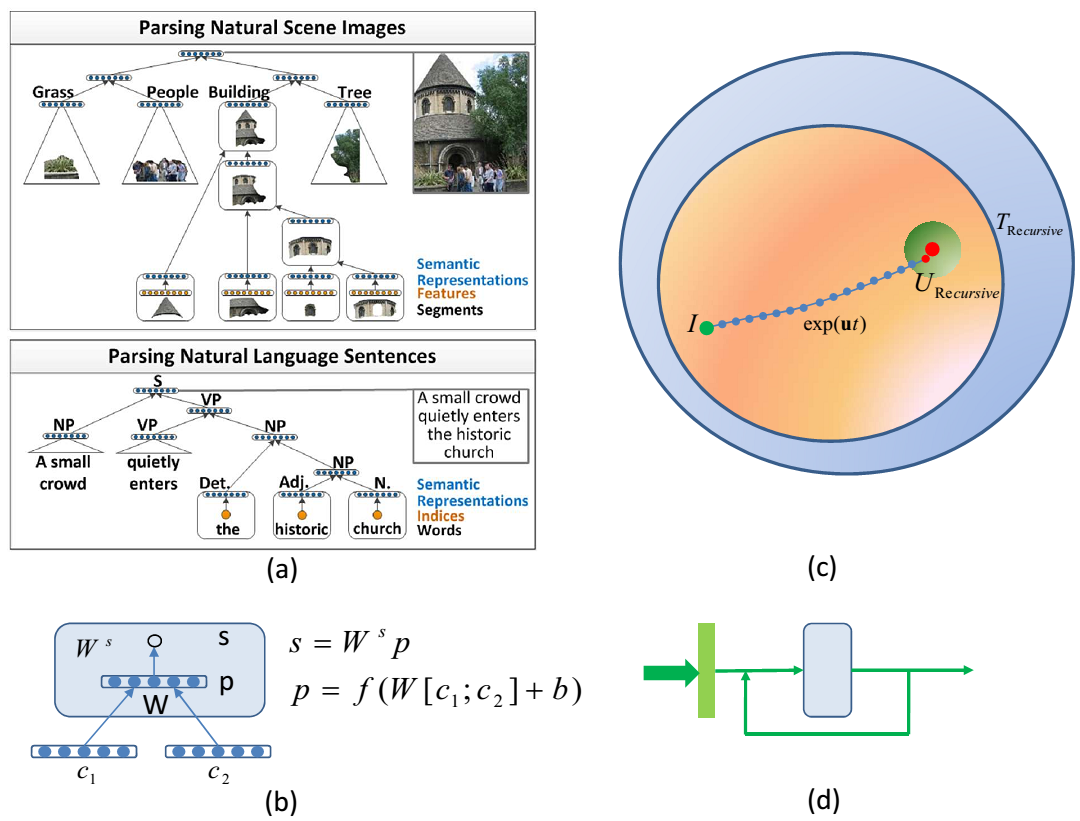}
  \caption{The geometry of recursive neural networks. (a) The recursive neural network architecture which parses images and natural language sentences \cite{Socher_Recursive}. (b)The fixed operation of a recursive neural network\cite{Socher_Recursive}. (c)The Lie group exponential mapping can be taken as the geometric picture of a recursive neural network. (d)The recursive neural network can optimize both the representation mapping and a component of the Lie algebra of the Lie group that can reach the destination oparation.}\label{fig6}
\end{figure}

It's easy to see that given the representation mapping and the adjacent matrix, the recursive operation is nothing but a transformation on $\tilde{\mathcal{X}}$. The fixed operation rule reminds us its similarity with the Lie group exponential mapping. Though we can not say $\tilde{\mathcal{X}}$ is a Lie group, but conceptually the recursive neural network can be understood as a discretized Lie group exponential mappings. Given any input in $\tilde{\mathcal{X}}$, the transformation between $\tilde{\mathcal{X}}$ and $\mathcal{Y}$ is achieved by this Lie group exponential mapping.

In CNN and ResNet, the goal is to find a \emph{free} curve with a fixed length, which is neither a geodesic nor a Lie group exponential mapping.
So the complexity or the dimension of freedom of the curve is much higher. In recursive neural networks, the exponential mapping like curve is much simpler, this is why we have the capability to optimize both the representation mapping and the exponential mapping simultaneously during training.

We can also observe that the recursive neural network does not emphasis on local operations as in CNNs or ResNets. This means that the recursive neural network is dealing a class of problems that is similar with the template matching problem. And its analogy with the SVF framework of the template matching confirms this point.

\subsection{Geometry of recurrent neural networks}
Another category of neural networks is the recurrent neural networks\cite{Lipton_RNN}. The key feature of the RNN is that it allows us to process sequential data and exploit the dependency among data. RNNs are widely used in language related tasks such as the language modeling, text generating, machine translation, speech recognition and image caption generation. Commonly used RNN structures include bi-directional RNNs, LSTM networks and deep RNNs.

Given a sequence of input data $\mathbf{x}=\{x_1,x_2,......,x_T\}$, a standard RNN compute the hidden vector sequence $\mathbf{h}=\{h_1,h_2,......,h_T\}$ and the output sequence $\mathbf{y}=\{y_1,y_2,......,y_T\}$ for every time step $t=1,2,...,T$ as:

\begin{eqnarray}\label{eq21}
  h_t &=& f(W_{ih}x_t+W_{hh}h_{t-1}+b_h) \\
  y_t &=& W_{ho}h_t+b_o
\end{eqnarray}
where $W_{ih},W_{hh},W_{ho}$ and $b_h,b_o$ are the weight matrices and the bias vectors. $f$ is the activation function of the hidden layer.

Here we are not going to give a complete overview to the different variations of RNNs. Instead we will try to understand the feature of RNNs from a geometrical point of view.

Let's first go back to the geometry of CNNs, where a CNN is a curve on $T_{CNN}$ to reach a point $U_{CNN}$. For each given data point on $V_{CNN}$, it's transformed by $U_{CNN}$ to another point on $V_{CNN}$. Geometrically a CNN accomplishes a point-to-point mapping with a simple curve as its trajectory on $V_{CNN}$.

RNN is more complex than CNN. We can check the geometric picture of RNNs in two different ways.

The first picture is to regard the sequential input data as a series of points on $V_{RNN}$, then the geometric picture of a RNN is a mapping of string, a series of interconnected points $\mathbf{x}=\{x_1,x_2,......,x_T\}$, to a destination string $\mathbf{y}=\{y_1,y_2,......,y_T\}$ with an intermediate trajectory $\mathbf{h}=\{h_1,h_2,......,h_T\}$. This is not a collection of independent parallel trajectories as $(x_1,h_1,y_1),(x_2,h_2,y_2),...,(x_T,h_T,y_T)$ since they are coupled by the recursive structure of the RNN through the hidden states $h_t$ in (\ref{eq21}).

Another picture of RNNs is to regard the sequential data, the string in the first picture, as a single data point, then the trajectory of the input data is a curve in a higher dimensional space $V_{RNN}$. Accordingly $T_{RNN}$ is the space of transformations on $V_{RNN}$. This is similar with the geometric picture of CNNs.

The difference of these two pictures is that in the first picture, the transformation working on the trajectory of each point of the string can be any transformation in $T_{RNN}$ and the trajectories of different points are coupled with each other. In the second picture, the transformation working on the data point, the complete sequence of data $\mathbf{x}={x_1,x_2,......,x_T}$, is now local operations in $T_{RNN}$, which only work on each individual data $x_i, i=1,2,...,T$ and pairs of successive data $x_i, x_{i+1}$ also through $h_t$ as in \ref{}. So the second picture is somehow trivial since now the RNN is only a special case of the CNN. So we from now on we will focus on the first picture to see how this picture can help us to understand the RNN related topics.

For a standard RNN, only earlier trajectories will influence later trajectories so that a later input data feels the \emph{force} of earlier data. For bi-directional RNNs, the \emph{force} is bi-directional. For deep bi-directional RNNs, the trajectory is a complex sheet consisting of multiple interconnected trajectories.

This string-sheet picture reminds us of their counterparts in physics. A deep CNN corresponds to the world-line of a particle, which can achieve human vision functions. A deep RNN corresponds to the world-sheet of a string, which can be used to solve linguistic problems. It's natural to ask, what's the correspondent structure of the trajectory of a membrane and what kind of brain function can be achieved by such a structure? Just as that RNNs can interconnect points into a string, we can also interconnect multiple strings to build a membrane. We can even further to interconnect multiple membranes to generate higher dimensional membranes.  To build such structures and investigate their potential functionalities may lead to interesting results.

We show that in fact the LSTM, the stacked LSTM, the attention mechanism of RNN and the grid LSTM are all special cases of this membrane structure.

\begin{figure}
  \centering
  \includegraphics[width=8cm]{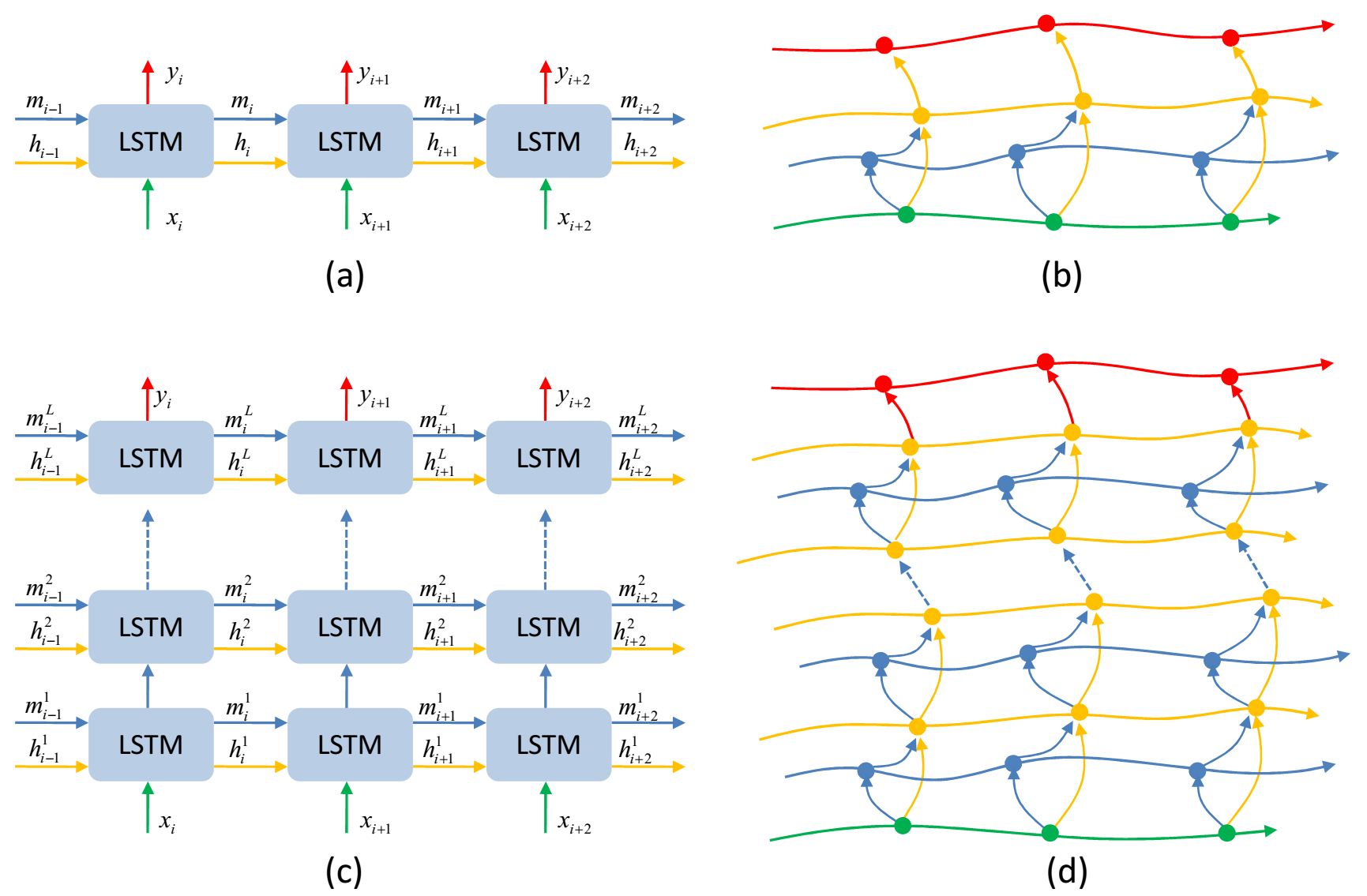}
  \caption{The geometry of LSTM networks. (a)The standard LSTM network where $m$ and $h$ are the memory and hidden states of the system. (b)The geometry of the LSTM network is given by two coupled strings, the yellow hidden state string and the blue memory string. (c) The structure of a stacked or deep LSTM network, whose geometric picture is a cascaded structure of coupled double strings as shown in (d). The arrows indicate information flows.}\label{fig-LSTM}
\end{figure}

\begin{figure}
  \centering
  \includegraphics[width=8cm]{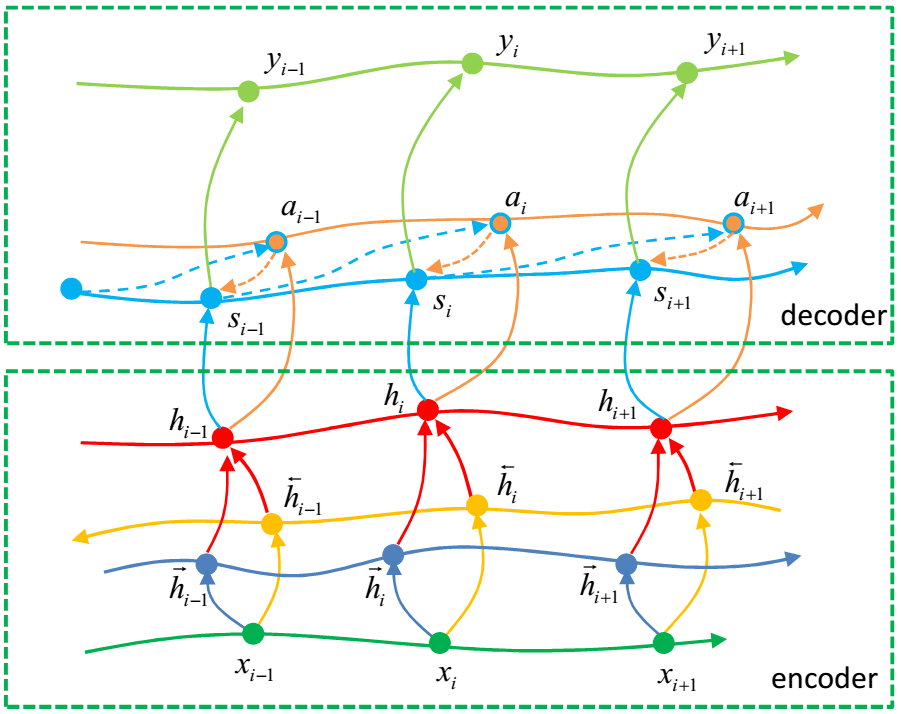}
  \caption{The geometry of the attention mechanism. We take the language translation system in \cite{Bahdanau_attentiontranslation} as an example. The encoder of this system is a bi-directional RNN network which generates encoded information ${h_i}$ from input series ${x_i}$. The decoder is a coupling of a RNN network (the light blue string) and an alignment model (the brown string) by exchanging state information ${s_i}$ of the RNN and the attention information ${a_i}$ from the alignment model, which generates the output ${y_i}$ from the encoded data ${h_i}$. This is a coupling of two heterogeneous networks. The arrows also indicate information flows}\label{fig-attention}
\end{figure}

\begin{figure}
  \centering
  \includegraphics[width=8cm]{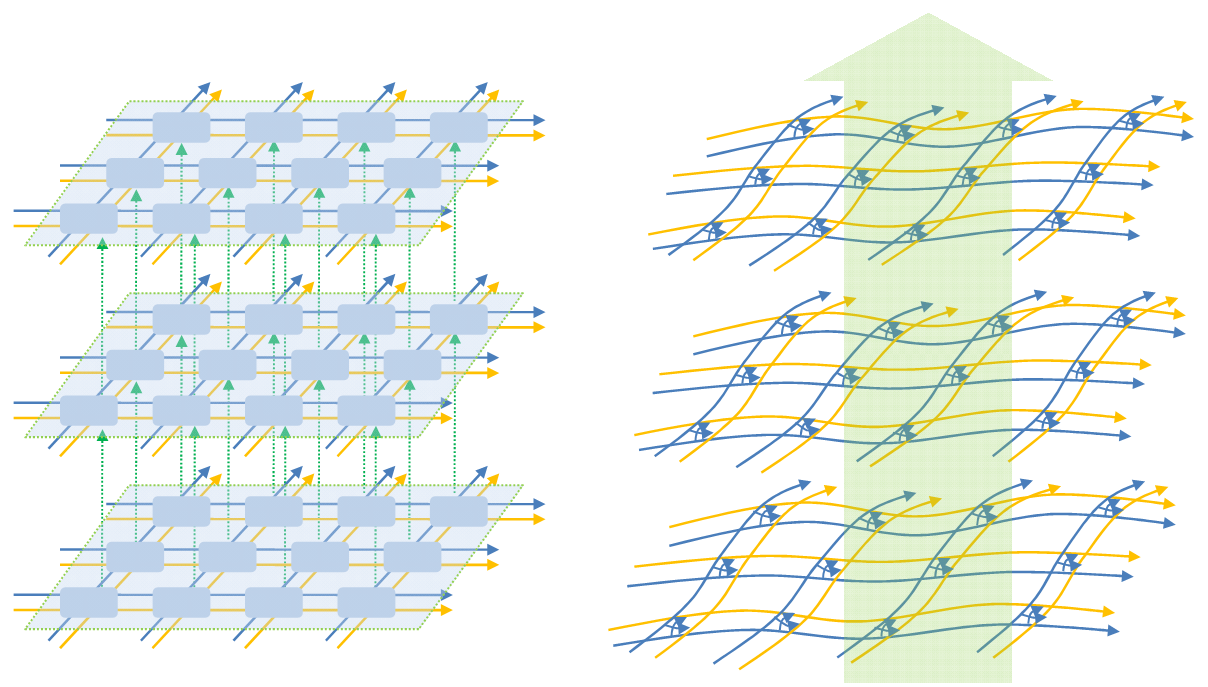}
  \caption{The geometry of the 2d deep grid LSTM. The 2d deep grid LSTM (left) has a geometric picture of multiple stacked membranes, where each membrane is constructed by coupling multiple LSTM strings in 2d space. }\label{fig-gridLSTM}
\end{figure}

\begin{itemize}

  \item \textbf{LSTM}\\ The structure of the standard LSTM is shown in Fig. \ref{fig-LSTM}. It's straight forward to see that it's the evolution of two parallel but coupled strings, the hidden states $h$ and the memory $m$. Compared with normal RNNs where there is only a single information passage, it's possible that the existence of the parallel information passages in LSTM makes the LSTM system capable of modeling long range dependencies.

  \item \textbf{Stacked LSTM}\\ The stacked LSTM is in fact a deep LSTM whose geometric picture is a cascaded double string as in Fig. \ref{fig-LSTM}.

  \item \textbf{Grid LSTM}\\ The grid LSTM\cite{Kalchbrenner2015_gridLSTM} is an extension of the LSTM to higher dimensions. Accordingly the strings in the standard LSTM are replaced by higher dimensional membranes as given in \cite{Kalchbrenner2015_gridLSTM}, where a 2d deep grid LSTM can be understood as stacked multiple 2d membranes. So if the deep LSTM is a cascaded coupled double strings, then the deep 2d grid LSTM is a cascaded membranes (Fig. \ref{fig-gridLSTM}). Similar geometric pictures can be constructed for higher dimensional and deep grid LSTM systems.

  \item \textbf{Attention mechanism}\\ The attention mechanism is getting popular in the decoder of the machine translation and the image caption generation. The key feature is that during the text generation, an alignment or an attention model (a feed-forward network in \cite{XuK2015Show}  or a multiple layer perceptron network in \cite{Bahdanau_attentiontranslation}) is used to generate a time-varying attention vector to indicate which part of the input data should be paid more attention to. This structure can be understood as a coupling between the normal decoder network (a RNN or a LSTM) with the attention network as shown in Fig. \ref{fig-attention}.
\end{itemize}

It need to be noted that there is an essential difference between the LSTM systems and the attention mechanism. We can see that the LSTM systems are the coupling of homogeneous subsystems but the attention mechanism is in fact a coupling of two heterogeneous subsystems, i.e. one decoder RNN/LSTM network and one attention model with a different structure. Other similar systems, such as the recurrent models of visual attention\cite{mnih2014_recurrentattention} and the neural Turing machines\cite{graves_neuralTM}\cite{Zaremba_neuralTM}, also hold a similar picture of coupling heterogeneous subnetworks.

It can be expected that more complex and powerful systems can be constructed by such a mechanism. It will be interesting to see how such a mechanism may help us to understand key human intelligence components such as our memory, imagination and creativity.

Similar with the case of CNNs, we can also check the metric and the curvature of the manifold of the coupled networks. Roughly we are now working with the direct product of the manifold of each subsystem with the tangent spaces of the subsystems as orthogonal subspaces of the tangent space of the complete manifold. This is very similar with the case of the semi-direct group product in the metamorphsis framework in the template matching system\cite{Holm2009_EP_Metamorphosis}. It can be expected that the structure of the manifold is much more complex than the relatively simple CNN case. But if the curvature of each subsystem's manifold is almost negative as in the quantum computation, then the convergence of the training of the coupled system naturally will be more difficult, just as indicated in the case of neural Turing machines.

\begin{figure}
  \centering
  \includegraphics[width=8cm]{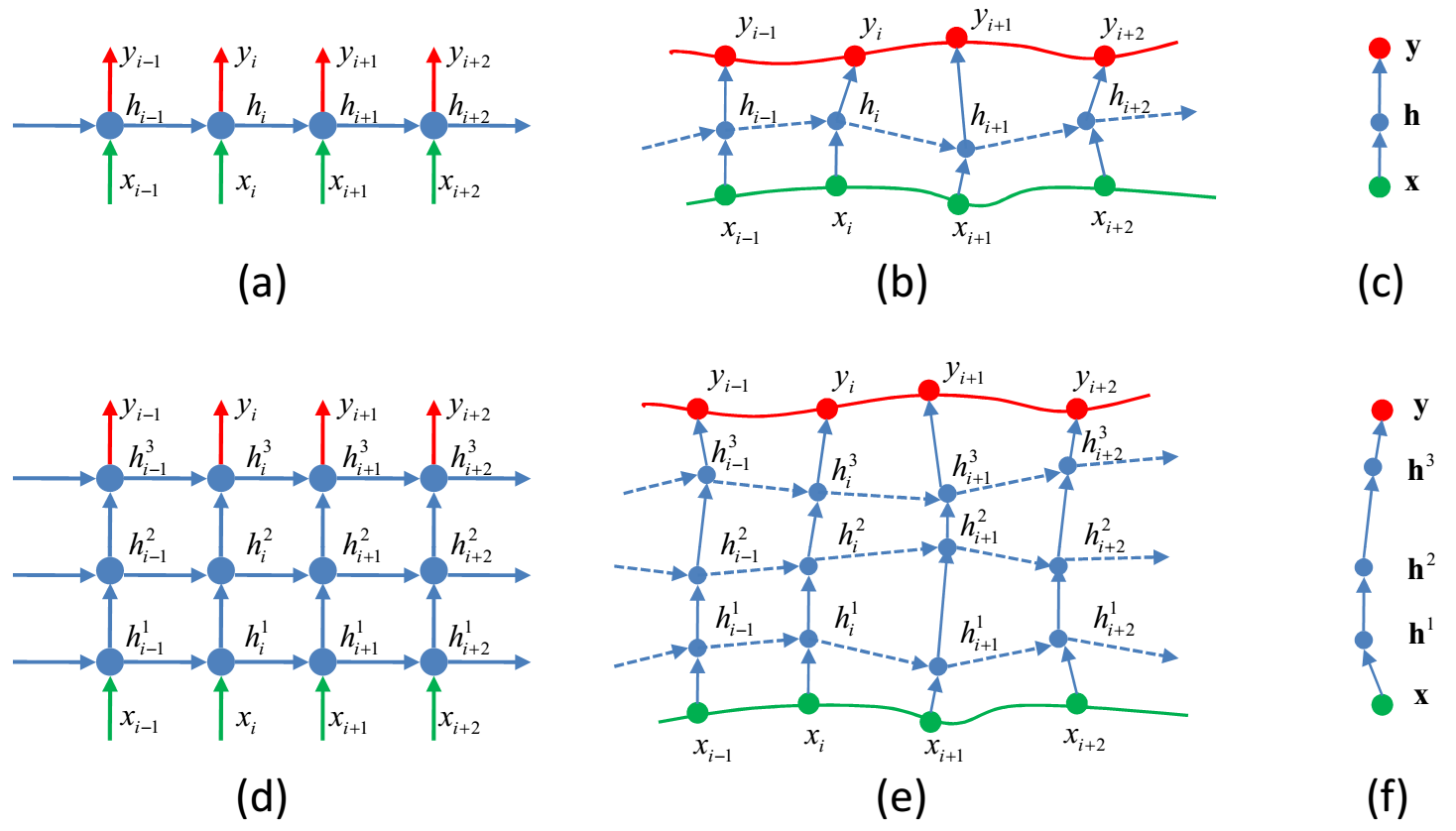}
  \caption{The geometry of RNNs. (a)The structure of the unrolled standard RNN. (b)The geometrical picture of the RNN is given by the trajectory of a string from the initial green string to the red destination string. This is the first way to understand the geometry of RNNs. (c) The second way to understand the geometry of RNNs is to take the input sequence as a single data point, then the structure of RNNs is a curve as the same as CNNs. The curve is now achieved by local operations on the data. (d) A deep RNN network with its correspondent geometric picture as a sheet swept by a string shown in (e) or a long curve shown in (f). The arrows indicate information flows}\label{fig-rnn}
\end{figure}

\subsection{Geometry of GANs}
Generative adversarial networks (GANs) is a framework for unsupervised learning. The key idea of GANs is to train a generative model that can approximate the real data distribution. This is achieved by the competition between a generative model $G$ and a discriminative model $D$. In the standard configuration of GANS, both $G$ and $D$ are CNN-like networks.

We now give a geometrical description of GANs and try to answer the following questions:
\begin{itemize}
  \item Why is it difficult to train GANs?
  \item Why can the refined GANs, including LAPGANs, GRANs and infoGANs, perform better than standard GANs?
\end{itemize}

\begin{figure}
  \centering
  \includegraphics[width=7cm]{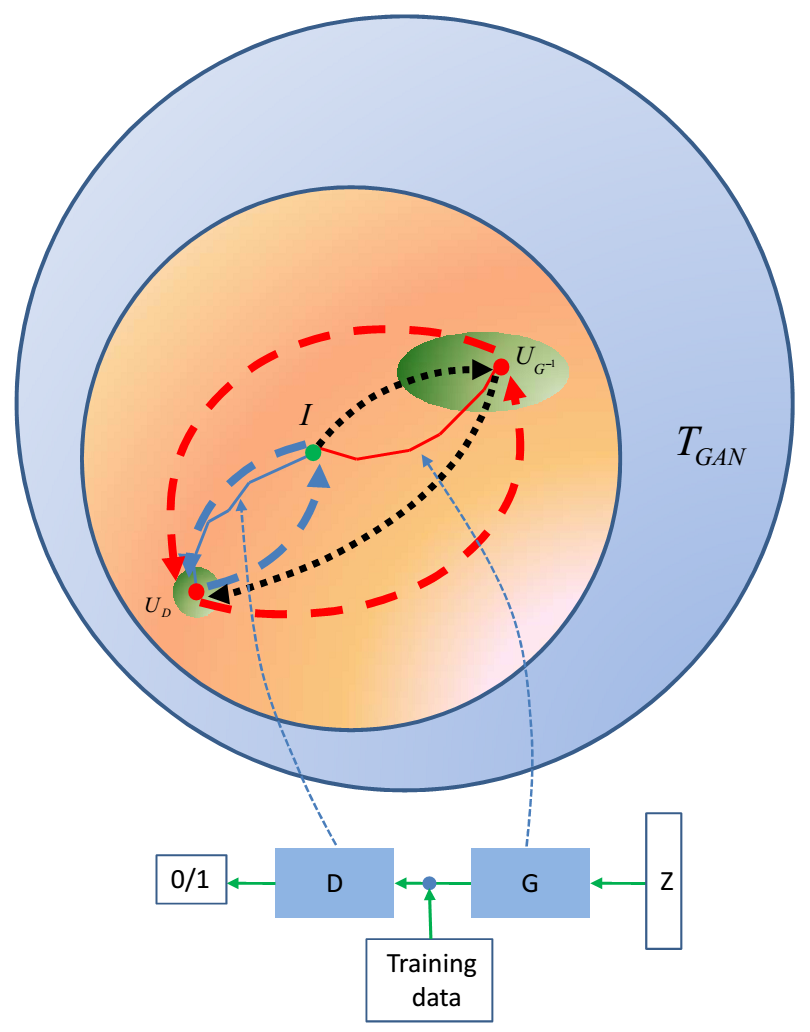}
  \caption{The geometry of GANs. The generator $G$ and discriminator $D$ of a GAN correspond to two curves connecting the identity operation $I$ with the transformations $U_{D}$ and $U_{G^{-1}}$ respectively. During the training procedure, the information flow (the feed-forward and the back-propagation of information) of D and G are shown by the blue and red dash lines. In the design of the GANs, the information flow is shown in the black dash line. The unsupervised learning of the generator G of GANs leads to a much flexible target transformation $U_{G^{-1}}$ compared with the target transformation $U_{CNN}$ in CNNs. }\label{fig-GAN}
\end{figure}

Since both $G$ and $D$ are realized by CNNs, the geometric picture of GANs can be built on the geometry of CNNs. From the geometric model of CNNs, it's easy to give the geometric picture of the standard DCGANs as a two-piece curve in $T_{CNN}$ as shown in Fig \ref{fig-GAN}.

\begin{figure}
  \centering
  \includegraphics[width=8.5cm]{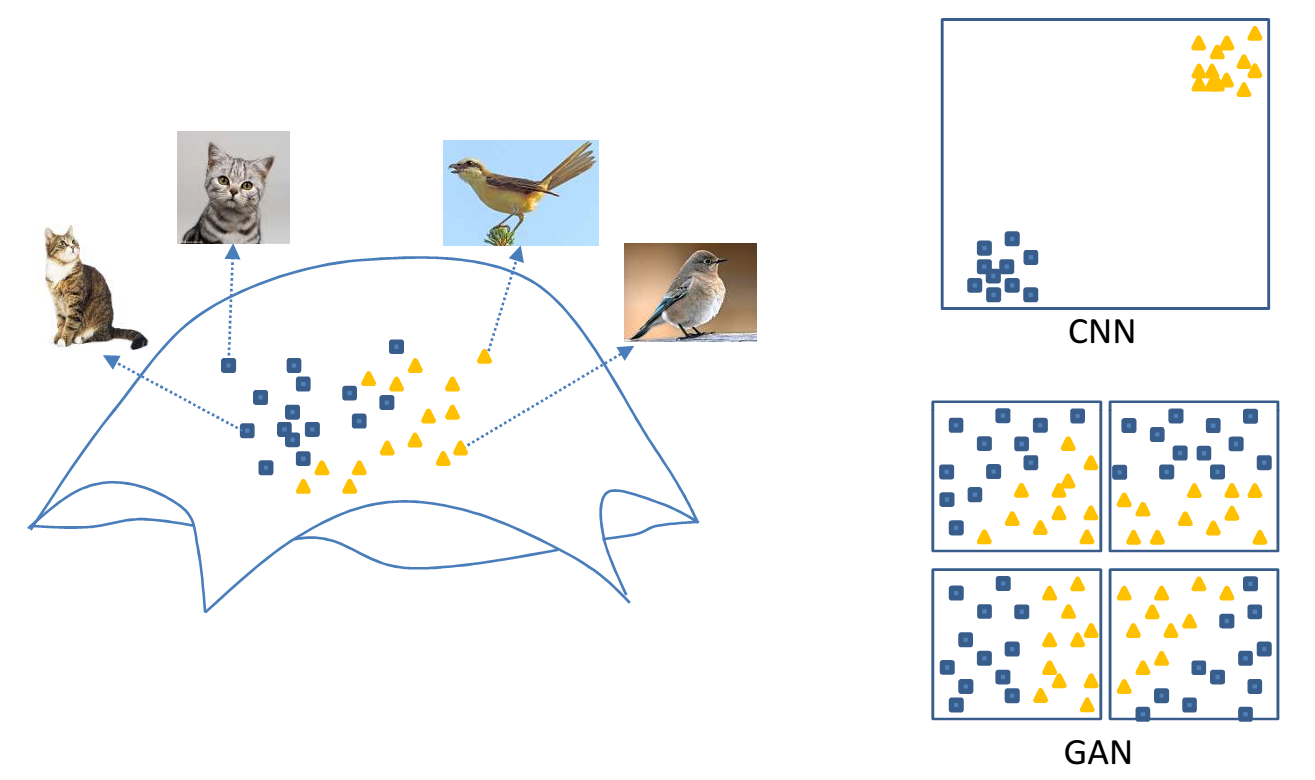}
  \caption{The comparison of the target transformation of CNNs and GANs. We consider a simple two-cluster case where the training data are images of cats and birds. Direct clustering and interpolation on the image manifold (left) is difficult. Both the CNN and the GAN map the training data into a 2 dimensional Euclidean space to achieve a proper data clustering (right). For the CNN, the output clustering pattern is unique. But for the GAN, we have multiple equivalent patterns that are all optima of the cost function of GAN systems.}\label{fig-CNNGAN}
\end{figure}

The two curve pieces represent the generator $G$ and the discriminator $D$ respectively. The goal of the training procedure is to find the two transformations that can minmax the cost function of GANs. GANs are different from CNNs in the following aspects:

\begin{itemize}
  \item The goal of CNNs is to find a curve $U_{CNN}(t)$ that can reach a transformation $U_{CNN}$ from $I$. For GANs, we need to find two transformations $U_{D}$ and $U_{G^{-1}}$ by constructing two curves to reach them from $I$ simultaneously.

  \item The target transformation of $D$ is relatively simple but the destination transformation of $G$ is much flexible than $U_{CNN}$. This can be easily understood since we set more constraints on $U_{CNN}$ by labeling the training data in CNN based object classification systems. An easy example to illustrate this point is shown in Fig. \ref{fig-CNNGAN}. Both CNNs and the inverse of G aim to transform the training data into an Euclidean space so that the training data are properly clustered. For CNNs, the labeling of training data results in that the output clustering pattern of CNNs is relatively unique. But for the unsupervised learning of the generator of GANs, their are a set of equivalent clustering patterns, which may compete with each other during the training procedure.

  \item The information pathways are also different in GANs and CNNs as shown in Fig. \ref{fig-GAN}. Firstly we can see that in the design of GANs, the basic idea is to learn a representation of the training data with $G^{-1}$ so that the generated examples from $G$ can have the same distribution of the training data, i.e. the generated data can fool the discriminator $D$. So the information flow in this design is to start from the training data, go through the reverse generator $G^{-1}$, return back through $G$ and pass the discriminator $D$. Only at this point, the output of $D$ can send error information back to guide the update structure of $G$. During the network training, the information flow for $D$ is similar with a normal CNN. But the optimal $G$ is defined by the cost function of GANs, which need an information passage passing both $G$ and $D$ to set constraints on $G$, which is longer than the CNN case.

  \item The geometric picture of GANs is to find a two-segment curve connecting the input space of $G$ to the output space of $D$, which is asked to pass the neighbourhood of the training data in a proper way. But we can only use the information from the output of $D$ to justify if the curve fulfills our requirement. So any loss of information at the output of $D$, due to either the mismatch between the training of $G$ and $D$ or an improper cost function that fails to measure the distance between the curve and the training data, will deteriorate the convergency of GANs.
\end{itemize}

Given the above observations, the reason that GANs are difficult to be trained is straight forward.
\begin{itemize}
  \item Compared with CNNs, GANs have the same Riemannian manifold as $T_{CNN}$, which has a high possibility of owning a negative sectional curvature. GANs have to optimize two curves and a more flexible target transformation $G$ on this manifold.
  \item The information pathways are longer in GANs than in CNNs. Also the information pathways in the training procedure do not coincide with the information pathways in the designing idea of GANs. This mismatch may lead to a deterioration of the system performance since the information flow in the $G^{-1}$ direction is lost in the training procedure.
\end{itemize}

To eliminate the difficulties of GANs, there exist the following three possible strategies:
\begin{itemize}
  \item To decrease the flexibility of the generator transformation $U_{G}$ by adding more constraints to $U_{G}$.
  \item To improve the structures of the curves of $G$ and $D$. From the analysis of CNNs, we know this is to change the metric of the manifold $T_{CNN}$ or equivalently to rearrange the lengths of the line segments in the curves of G and D.
  \item To eliminate the information loss at the output of $D$ so that it's more reliable to justify how the two-segment curve of GAN passes the neighbourhood of the training data. Basically this is exactly what the WGAN does.
  \item To change the information pathways during the training procedure. This may improve the efficiency of the information flow during the training so that possibly a better convergence may happen.
\end{itemize}

\begin{figure}
  \centering
  \includegraphics[width=8.5cm]{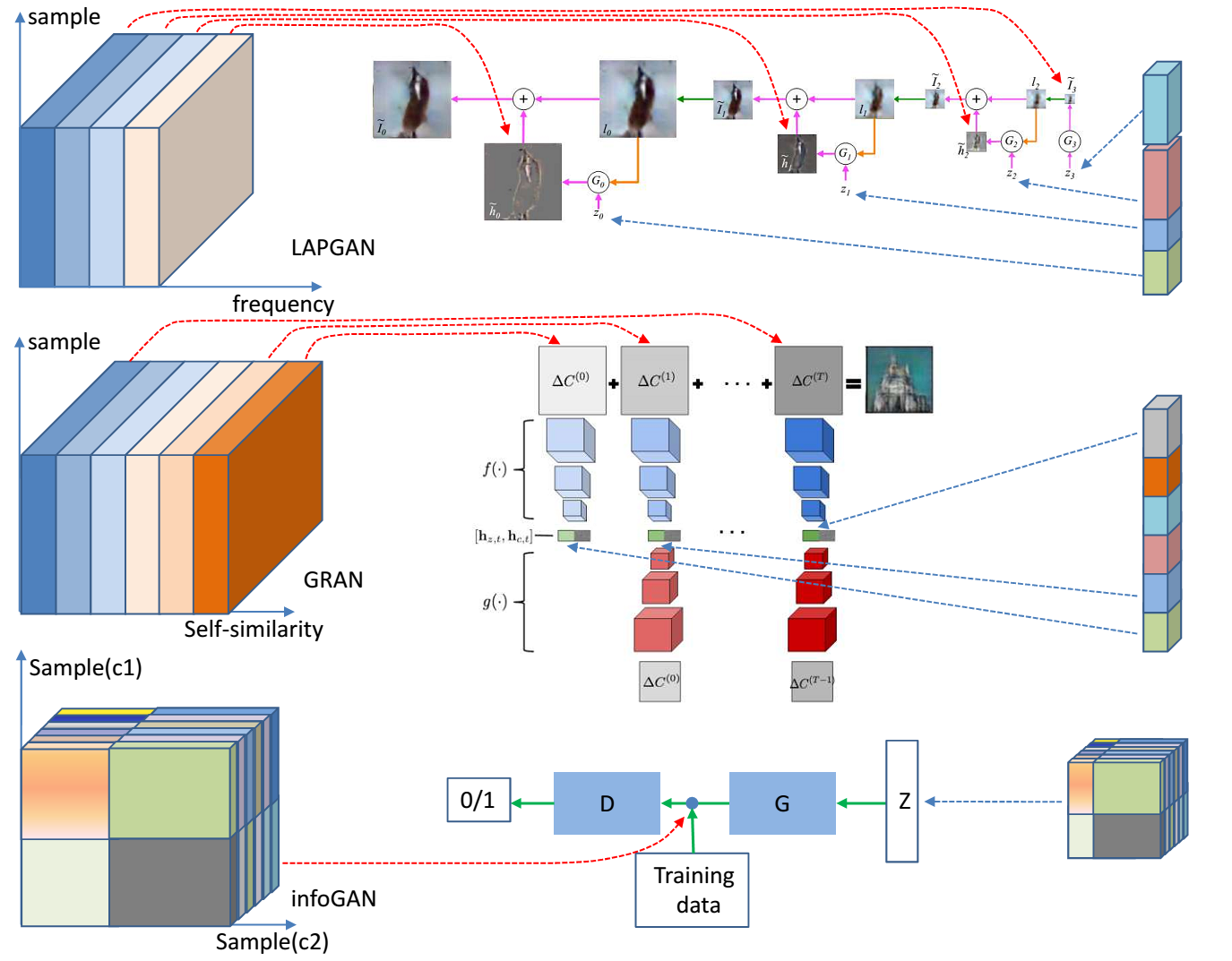}
  \caption{Comparisons of the hierarchical structures in LAPGANs,GRANs and infoGANs. Left column: The training data space; Central column: The structure of different GAN systems; Right column: The representation space at the input of generators. LAPGANs: The hierarchical structure is built on the frequency bands of all the training data and a corresponding structure in the representation space is also constructed by LAPGAN's pyramid structure. GRANs: The hierarchical structure is a kind of self-similarity or a recursive structure on the training data space encoded in the RNN based generator. infoGANs: The hierarchical structure is based on the clustering of the training data, where the clustering are indicated by the latent variables ($\{c_i\}$) in the generator's input data space, so that this cluster based structure is kept by the generator and can be verified by the mutual information used in infoGANs. In each system, a hierarchical correspondence between the training data space and the representation space is built, which can effectively introduce more constraints on the transformation $U_{G}$ so that the convergence of the training procedure can be improved. }\label{fig-GANS}
\end{figure}

Now we explain how LAPGANs, GRANs and infoGANs can improve the performance of GANs utilizing the above given strategies.
\begin{itemize}

  \item \textbf{InfoGAN}\\InfoGANs aim to decrease the flexibility of the transformation $U_{G^{-1}}$ or equivalently $U_{G}$ by introducing extra constraints on $U_{G}$. The original GAN cost function is based on the sample based expectation of the network outputs. On the contrary, infoGANs enhance the constraints on $U_{G}$ by setting constraints on expectations of clusters of sample data. This means that the clustering pattern in the representation space of the training data, indicated by the latent variables in the input data space of the generator $G$, should be kept by the generator, so that in the hidden states of the discriminator (as a kind of decoder of the generator to reveal the clustering pattern) the same clustering pattern can be observed.  This is exactly why the infoGAN add a mutual information item, which is used to check if the clustering patten, to its cost function. This cluster based constraints in fact constructs a hierarchical structure in both the training data space and the representation space. Asking the generator G to keep this hierarchical structure effectively reduce the flexibility of $U_{G}$ and improves the convergence stability.

  \item \textbf{LAPGAN}\\ The key feature of LAPGANs is to decompose the image space into orthogonal subspaces with different frequency bands and run different GANs sequentially on subspaces at higher frequency bands conditioned on the results on lower frequency bands. From our geometrical point of view, the decomposition of input data space results in a correspondent decomposition of the transformation space $T_{GAN}$ into multiple subspaces $T_{GAN}^{k}$ where $k=1,2,...,K$ with $K$ the number of bands of the paramid.  So the pyramid based LAPGAN is essentially to construct the curve $U_{D}(t)$ and $U_{G^{-1}}(t)$ in $T_{GAN}$ by a set of curves in each subspace $T_{GAN}^{k}$ using simpler local operations. This is similar to achieving a displacement in 3 dimensional space by three displacements in x, y and z directions.

      The reason such a structure can improve the convergence of LAPGANs can be understood from different points of view.

      Firstly LAPGANs replace a complex curve $U_{G}(t)$ by a set of simpler curves $U_{G}^{k}(t)$ in each orthogonal subspace $T_{GAN}^{k}$ and then optimize those curves sequentially. So the local operations constructing the curve $U_{G}^{k}(t)$  are simpler and the curve $U_{G}^{k}(t)$ is also shorter than $U_{G}(t)$.
      From our above analysis of CNNs and ResNets, simpler local operations will introduce a flatter transformation manifold or equivalently a curve consisted of shorter line segments. This will improve the convergence property of each curve $U_{G}^{k}$. This is to say a curve $U_{G}(t)$ with complex local operations by a series of shorter curves $\{U_{G}^{k}\}$ with simpler local operations in subspaces $\{T_{GAN}^{k}\}$.

      Secondly in LAPGANs, a set of generators $\{G^{k}\}$ and discriminators $\{D^k\}$ are trained with each subband of training data. Obviously each pair of $G^k$ and $D^k$ determines a representation pattern of one data subband. Also these representations are consistent since during the training procedure, the generator $G^k$ at a higher frequency band is conditioned on the lower band image. This means LAPGANs replace the original cost function on the expectation of the complete data set by a set of cost functions on the expectation of each frequency band of the input data. This is in fact to construct a hierarchical structure on the training data, i.e. the paramid, and set a set of consistent constraints on this hierarchical structure. So essentially we add more constraints to the generator transformation $U_{G}$.

  \item \textbf{GRAN}\\ GRANs improve the performance of GANs by a RNN based generator so that the final generated data is constructed from the sequential output of the RNN network. We already know that the RNN essentially decompose a single input data point of a normal CNN network into a data sequence and therefore the trajectory of a data point of CNNs becomes a sheet swept by a string in RNNs. From this point of view, the operations of a RNN are essentially all simpler local operations of the complete input data. What's more wimilar with LAPGANs, GRANs also explore a hierarchical structure of the system which is encoded in the RNN structure. But this hierarchical structure is not so explicit as in LAPGANs. In another word, LAPGANs explore the structure in frequency domain and GRANs explore a kind of self-similarity like structure of the input data.

\end{itemize}

As a summary, LAPGANs, GRANs and infoGANs all improve the performance by constructing a hierarchical structure in the system and adding more constraints on $U_{G}$. Their difference lies in their ways to construct the hierarchy. What's more, LAPGANs and GRANs can potentially further improve the performance by changing the structure of the curve $U_{G}(t)$ by only using simpler local operations compared with the standard GANs.

Another possible way to improve the performance is to change the information passway. For example in the training of $G$ and $D$, the training data information is fed to D directly but
to G indirectly. Feeding the training data directly to G in the inverse direction (as the information passway given by the black dash arrow in Fig. \ref{fig-GAN}) can potentially further explore the structure of the representation space $z$.

\subsection{Geometry of equilibrium propagation}
Equilibrium propagation\cite{Scellier_equilibrium} is a new framework for machine learning, where the prediction and the objective function are defined implicitly through an energy function of the data and the parameters of the model, rather than explicitly as in a feedforward net. The energy function $F(\theta,\beta,s,v)$ is defined to model all interactions within the system and the actions with the external world of the system, where $\theta$ is the parameter to be learned, $\beta$ is a parameter to control the level of the influence of the external world, $v$ is the state of the external world(input and expected output) and $s$ is the state of the system.

The cost function is define by

\begin{equation}\label{eq3}
  C_{\beta}^{\delta}(\theta,s,v):=\delta^{T}\cdot\frac{\partial F}{\partial \beta}(\theta,\beta,s_{\theta,v}^{\beta},v)
\end{equation}

where $\delta$ is a directional vector in the space of $\beta$ so the cost function is the directional derivative of the function $\beta\mapsto F(\theta,\beta,s,v)$ at the point $\beta$ in the direction $\delta$.

For fixed $\theta$,$\beta$ and $v$, a local mininum $s_{\theta,v}^{\beta}$ of $F$ which corresponds to the prediction from the model, is given by

\begin{equation}\label{eq4}
  \frac{\partial F}{\partial s}(\theta,\beta,s_{\theta,v}^{\beta},v)=0
\end{equation}

Then the equilibrium propagation is a two-phase procedure as follows:
\begin{enumerate}
  \item Run a 0-phase until the system settle to a 0-fixed poin $s_{\theta,v}^{\beta}$ and collect the statistics $\frac{\partial F}{\partial \theta}(\theta,\beta,s_{\theta,v}^{\beta},v)$.
  \item Run a $\xi$-phase for some small $\xi\neq 0$ to a $\xi$-fixed poin $s_{\theta,v}^{\beta+\xi\delta}$ and collect the statistics $\frac{\partial F}{\partial \theta}(\theta,\beta,s_{\theta,v}^{\beta+\xi\delta},v)$.
  \item Update the parameter $\theta$ by
  \begin{equation}\label{eq5}
    \Delta\theta \propto -\frac{1}{\xi}(\frac{\partial F}{\partial \theta}(\theta,\beta,s_{\theta,v}^{\beta+\xi\delta},v)-\frac{\partial F}{\partial \theta}(\theta,\beta,s_{\theta,v}^{\beta},v)).
  \end{equation}
\end{enumerate}

An equivalent constrained optimization formulation of the problem is to find
\begin{equation}\label{eq6}
  \arg_{\theta,s}\min C_{\theta,s}^{\beta}(\theta,s,v)
\end{equation}
 subject to the constrain
 \begin{equation}\label{eq7}
   \frac{\partial F}{\partial s}(\theta,\beta,s_{\theta,v}^{\beta},v)=0.
 \end{equation}
This leads to the Lagrangian
\begin{equation}\label{eq8}
  L(\theta,s,\lambda)=\delta\cdot\frac{\partial F}{\partial \beta}(\theta,\beta,s_{\theta,v}^{\beta+\xi},v)+\lambda\cdot\frac{\partial F}{\partial s}(\theta,\beta,s_{\theta,v}^{\beta},v)
\end{equation}

By this formulation we can see the update of $\theta$ in equilibrium propagation is just one step of gradient descent on $L$ with respect to $\theta$ as
\begin{equation}\label{eq9}
  \Delta\theta\propto -\frac{\partial L}{\partial \theta}(\theta,s^{*},\lambda^{*})
\end{equation}
where $s^{*},\lambda^{*}$ is determined by
\begin{equation}\label{eq10}
  \frac{\partial L}{\partial \lambda}(\theta,s^{*},\lambda^{*})=0
\end{equation}
and
\begin{equation}\label{eq11}
  \frac{\partial L}{\partial s}(\theta,s^{*},\lambda^{*})=0
\end{equation}

It's claimed that the $\xi$-phase can be understood as performing the back-propagation of errors. Because when a small $\xi\delta$ on $\beta$ is introduced, the energy function $F$ induces a new 'external force' acting on the output units of the system to drive the output to output target. This perturbation at the output layer will propagates backward across the hidden layers of the network so that it can be thought of a back-propagation.

Here we will show that the geodesic shooting algorithm\cite{Vialard2011_GS_Adjoint}\cite{Vialard2012_GS_Atlas} in the diffeomorphic template matching is in fact an example of the equilibrium propagation. We will use it to give a geometrically intuitive picture of the equilibrium propagation to show how the back-propagation is achieved in the $\xi$-phase.

The geodesic shooting addresses that diffeomorphic template matching as find the optimal matching $g(t),\dot{g}(t)=u(t)\circ g(t)$ between two images $I_{0}$ and $I_{1}$. Under the smooth transformation of $g(t)$, we define the transformed source image as $I(t)=I_{0}\circ g(t)$. The energy function is given as

\begin{equation}\label{eq15}
  F=\int_0^1 \langle u(t),u(t)\rangle_v dt+\beta\|I_1-I_0\circ g_{1}\|_{L^2}^2
\end{equation}

In the language of the equilibrium propagation, $v$ corresponds $I_0$ and $I_1$ as the input and output data, the state of the system $s$ is given by ${I(t),g(t),u(t)}$). The cost function is $C_{\theta,s}^{\beta}(\theta,s,v)=\delta\|I_1-I_0\circ g_{1}\|_{L^2}^2$.The prediction of the state $s^{*}$ is obtained by $\frac{\partial F}{\partial s}=0$, which leads to the Euler-Pointcare equation given by
\begin{eqnarray}
  \dot{I}(t) &=& -\hat{P}(0)I(t)\cdot u(t) \\
  \dot{\hat{I}}(t) &=& -\hat{P}(0)\cdot(\hat{I}(t)u(t)) \\
  Lu(t) &=& --\hat{P}(0) I(t)\cdot \hat{I}(t) \\
  \hat{I}(1) &=& -(I(1)-I_1)
\end{eqnarray}
where $Lu(t)$ is the momentum map and $\hat{I}(t)$ is the adjoint variable of $I(t)$. It's easy to know that the optimal state $s^{*}$ is completely determined by $u(0)$, which can be regarded as the parameter $\theta$ to be optimized.

Taking the EP equation as the constraint, the variation of \ref{eq8} results in

\begin{eqnarray}
  \dot{I}(t) &=& -\nabla I(t)\cdot u(t) \\
  \dot{P}(t) &=& -\nabla\cdot(P(t)u(t)) \\
  Lu(t) &=& --\hat{P}(0) I(t)\cdot P(t) \\
  \dot{\hat{I}}(t) &=& -\nabla\cdot(\hat{I}(t)u(t))-\nabla\cdot((K*\hat{u}(t)P(t)))\\
  \dot{\hat{P}}(t) &=& -\nabla\hat{P}(t)\cdot u(t)+(K*\hat{I}(t))\cdot \nabla I(t)\\
  K*\hat{I}(t)&=&-u(t)-K*(\hat{I}(t)\nabla I(t)-\nabla\hat{P}(t)P(t))\\
  \hat{I}(1)&=& -(I(1)-I_1)\\
  \hat{P}(1) &=& 0\\
  -\hat{P}(0)&=&\nabla_{P(0)}C_{\theta,s}^{\beta}(\theta,s,v)
\end{eqnarray}

where $P(t)=Lu(t)$ is the momentum map of $u(t)$.

In the geodesic shooting framework, the above equations include two components:
\begin{itemize}
  \item \textbf{Shooting} The first three equations determine $I(t),P(t)$ given the initial momentum $P(0)$ and $I(0)=I_{0}$.
  \item \textbf{Adjoint advection}\\ The second five equations determines $\hat{I}(t),\hat{P}(t)$ by solving the equations in the reverse time direction given the boundary conditions of $\hat{I}(1),\hat{P}(1)$.
  \item \textbf{Gradient}\\ The last equation given the gradient of the cost function $C_{\theta,s}^{\beta}(\theta,s,v)$ on the initial momentum $P(0)$, while updating $P(0)$ equals to update the parameter $\theta=u(0)$.
\end{itemize}

It's very easy to see that the three components correspond exactly to the three step in the equilibrium propagation. The shooting and advection phases correspond to the 0-phase and $\xi$-phase. The gradient $\nabla{P(0)}C_{\theta,s}^{\beta}(\theta,s,v)$ is equivalent to the gradient descent $\Delta\theta\propto -\frac{\partial L}{\partial \theta}(\theta,s^{*},\lambda^{*})$.

This example gives an intuitive picture of the equilibrium propagation as follows:
\begin{itemize}
  \item The 0-phase is to shoot a geodesic staring from a given point with a given initial velocity vector.
  \item The $\xi$-phase is to propagate the error information (the distance between the end point of the geodesic $I(1)$ and the expected destination $I_{1}$) in a backward direction.
  \item The update of parameter $\theta$ is based on the information $-\hat{P}(0)$ obtained by the back-propagation procedure.
\end{itemize}

Of course, the equilibrium propagation is a general framework for deep learning. We can not give a concrete geometric description of it due to its flexibility. Here we just show that in some special cases, the equilibrium propagation does have a geometric picture, in which how the back-propagation of error information can be directly observed.

Another observation is that the equilibrium propagation is not only a framework for training the so called \emph{implicitly} defined deep learning systems as described in \cite{Scellier_equilibrium}, in fact it's also potentially a framework to optimize the structure of the explicitly defined deep learning systems as well.

To see this point, taking the deep CNN as an example where the prediction is explicitly defined by the structure of the CNNs. We can define an energy as

\begin{equation}\label{eq16}
  F(\theta,\beta,s,v)=E_{comp}(\theta,\beta,v)+E_{CNN}(\theta,\beta,s,v)
\end{equation}
where the parameter $\theta$ to be learned includes also the parameters of the structure of a CNN, for example the depth, the number and size of each layer, besides the normal papameters of a CNN.  $E_{CNN}(\theta,\beta,s,v)$ is then the cost function for a CNN with a given structure defined by $\theta$ and $E_meta(\theta,\beta)$ defines the complexity of the CNN, i.e., the number of operations of the feed-forward network structure defined by $\theta,\beta$. $E_{comp}(\theta,\beta,v)$ does not depend on $s$ since the state of the CNN $s$ is determined by $\theta$ and $v$.

If we run the equilibrium propagation on this system, the system prediction $s^{*}$ can be achieved by the current gradient descent based CNN training method, i.e., training a CNN with a fixed configuration. Then the optimization on $\theta$ is in fact a global optimization framework to find the optimal CNN structure defined by $\theta^{*}$. Geometrically this is to find a Riemannian manifold $T_{CNN^{\theta^*}}$ with its metric $g_{CNN}^{theta^*}$ defined by $\theta^*$, so that the optimal CNN can realize a geodesic $U_{CNN}^{geo}(t)$, which gives a minimal system complexity, i.e., the most efficient CNN structure that can achieve the task.

Of course, the discrete parameters of the structure of the CNN can not be updated by the gradient descent strategy. This is the same as in the reinforcement learning neural Turing machines\cite{graves_neuralTM} if we regard the discrete structure parameters as the actions in the neural Turing machines. So theoretically we can train the systems also with the reinforcement learning to find the optimal structure parameters, while the back-propagation in \cite{graves_neuralTM} is replaced by the equilibrium propagation. Also this framework might not work due to the complexity in practice, just as the neural Turing machine can only achieve simple algorithmic tasks.

\section{Conclusional remarks}

This paper is based on the belief that geometry plays a central role in understanding physical systems. Based on a brief overview of the geometric structures of two fields, the quantum computation and the diffeomorphic computational anatomy, we show that deep learning systems hold a similar geometric picture and the properties of deep learning systems can be understood from this geometric perspective.

We summarize our observations on different deep learning systems as follows:

\textbf{Convolutional neural networks}
\begin{itemize}
  \item The structure of a CNN defines a manifold, the transformation space $T_{CNN}$, and the Riemannian metric on it.
  \item The goal of a CNN is to find a transformation $U_{CNN}$ on this Riemannian manifold, which is achieved by a curve $U_{CNN}(t)$ defined by the local operations of the CNN structure.
  \item A CNN is not to find the geodesic but to find a curve with a given structure to reach a point in the Riemannian manifold. A deep CNN works only when the curve it defines is longer than the geodesic between the identity operator $I$ and $U_{CNN}$. A too shallow network will never work efficiently.
  \item The convergence property of a CNN depends on both the curvature of the Riemannian manifold and the complexity distribution of the curve. The linear complexity defined by the configuration of kernels and the nonlinear complexity defined by the activation function should be balanced. A too deep network with improper complexity distribution may also fail.
  \item Generally to find the optimal CNN structure is difficult.
  \item How the structure of a CNN influence the curvature of the manifold and the performance of the network needs to be investigated.
  \item An alternative geometric structure is built on the data space $V_{CNN}$, where a CNN works by disentangling a highly curved input manifold into a flatter output manifold.

\end{itemize}

\textbf{Residual networks}
  \begin{itemize}
    \item Residual networks is a special case of CNNs, where the curve to reach the target transformation consists of a large number of short line segments.
    \item Each line segment corresponds to a near-identity transformation on the manifold, which can be achieved by the one-order approximation. The optimal ResNet structure is determined by this one-order approximation.
     \item The superior performance of ResNet comes from the balanced complexity distribution along the curve and a potentially smaller parameter space defined by the ResNet structure.
  \end{itemize}

\textbf{Recursive neural networks}
  \begin{itemize}
    \item Recursive neural networks have a structure that can be understood as a Lie group exponential mapping if the manifold of transforamtions is regarded as a Lie group.
    \item The relative simple structure of the recursive neural network makes it possible to find the representation mapping and the recursive operation during the training procedure.
  \end{itemize}

\textbf{Recurrent neural networks}
\begin{itemize}
  \item Recurrent neural networks can be understood as a sheet swept by a string on the data space $V_{RNN}$.
  \item A stacked or deep LSTM network accomplishes a cascaded structure of a pair of coupled strings.
  \item The grid LSTM network extends the coupled stings into weaved membranes.
  \item The attention mechanism and the neural Turing machine are coupled structure of heterogeneous subnetworks.
\end{itemize}

\textbf{Generative adversarial networks}
\begin{itemize}
  \item Generative adversarial networks are the simultaneous optimization of two curves leading to $U_G$ and $U_D$, in $T_{GAN}$, where each curve has a similar structure as CNNs. The unsupervised learning of the target transformation $U_G$ is much flexible than than the supervised learning of $U_{CNN}$. These features explain the difficulty of training GANs.
  \item LAPGANs achieve a better performance by both exploring the frequency domain structure of the training data and constructing the curve $U_{G}$ by simpler local operations on subspaces of the training data.
  \item GRANs also work in a similar way as LAPGAN. The difference is that GRANs construct a self-similarity like structure which is encoded in the RNN structure of $G$. Also GRANs only apply simpler local operations in $T_{GAN}$.
  \item InfoGANs reduce the flexibility of $U_G$ by setting extra constraints on clusters of data samples.
  \end{itemize}

\textbf{Equilibrium propagation}
  \begin{itemize}
    \item The geodesic shooting algorithm of the template matching is an example of the equilibrium propagation. It provides an intuitive picture of how the equilibrium propagation can fill the gap between the \emph{implicit} framework of optimization of a cost function and the back-propagation of information in the \emph{explicit} system.
    \item Theoretically the equilibrium propagation can be used as a framework for find the optimal deep learning system.
  \end{itemize}

\bibliographystyle{ieeetr}
\bibliography{bibfile}





\end{document}